\documentclass{article}

\usepackage[final]{neurips_2023}

\usepackage[utf8]{inputenc} % allow utf-8 input
\usepackage[T1]{fontenc}    % use 8-bit T1 fonts
\usepackage{hyperref}       % hyperlinks
\usepackage{url}            % simple URL typesetting
\usepackage{booktabs}       % professional-quality tables
\usepackage{amsfonts}       % blackboard math symbols
\usepackage{nicefrac}       % compact symbols for 1/2, etc.
\usepackage{microtype}      % microtypography
\usepackage{xcolor}         % colors

\usepackage{graphicx}
\usepackage{bbold}
\usepackage{bm}
\usepackage{amsmath,amscd,amssymb}
\usepackage{colortbl}

\usepackage{thm-restate}

\usepackage{scalerel}
\usepackage{stackengine,wasysym}
\usepackage{enumitem}

\usepackage{algorithm, algorithmicx, algpseudocode}

\newcommand\reallywidetilde[1]{\ThisStyle{%
  \setbox0=\hbox{$\SavedStyle#1$}%
  \stackengine{-.1\LMpt}{$\SavedStyle#1$}{%
    \stretchto{\scaleto{\SavedStyle\mkern.2mu\AC}{.5150\wd0}}{.6\ht0}%
  }{O}{c}{F}{T}{S}%
}}

\DeclareMathAlphabet{\mathscr}{U}{mathc}{m}{it}

\newcommand{\figref}[1]{Fig.~\ref{#1}}

\newcommand{\bz}{\mathbf{z}}

\newcommand{\bx}{\mathbf{x}}

\newcommand{\EE}{\mathbb{E}}
\newcommand{\elbo}{{\rm ELBO}}

\newcommand{\NN}{\mathcal{N}}

\newcommand{\ZZ}{\mathcal{Z}}

\newcommand{\AAA}{\mathcal{A}}
\newcommand{\AAAbar}{\overline{\mathcal{A}}}

\newcommand{\RR}{\mathbb{R}}

\newcommand{\KL}{{\rm KL}}

\newcommand{\bth}{{\bm{\theta}}}
\newcommand{\bphi}{{\bm{\phi}}}

\newcommand{\LL}{\mathcal{L}}

\newcommand{\target}{{\rm tgt}}
\newcommand{\pth}{p_{\bth}}
\newcommand{\qphi}{q_{\bphi}}
\newcommand{\qphin}{q_{\bphi_n}}
\newcommand{\rr}{r_{\bphi, \bth}}
\newcommand{\rreps}{r_{\bphi, \bth, \epsilon}}
\newcommand{\rrn}{r_{\bphi_n, \bth}}
\newcommand{\aaa}{a_{\bphi, \bth}}
\newcommand{\aaaeps}{a_{\bphi, \bth, \epsilon}}
\newcommand{\aaan}{a_{\bphi_n, \bth}}
\newcommand{\ZZr}{\ZZ_r}
\newcommand{\ZZp}{\ZZ_p}
\newcommand{\ellT}{\ell_{\bth, \bphi}^T}

\newcommand{\secref}[1]{Sec.~\ref{#1}}
\newcommand{\eqnref}[1]{Eqn.~\ref{#1}}

\usepackage{todonotes}
\setuptodonotes{inline}

\title{Reparameterized Variational Rejection Sampling}

\author{%
  Martin Jankowiak \\ %\thanks{}
  Generate Biomedicines \\
  Somerville, MA, USA \\
  \texttt{mjankowiak@generatebiomedicines.com} \\
  \And
  Du Phan \\ %\thanks{}
  Google Research \\ 
  Cambridge, MA, USA \\
  \texttt{phandu@google.com} \\
}

\begin{document}

\maketitle

\begin{abstract}
Traditional approaches to variational inference rely on parametric families of variational distributions,
with the choice of family playing a critical role in determining the accuracy of the resulting posterior approximation.
Simple mean-field families often lead to poor approximations, while rich families of distributions like normalizing
flows can be difficult to optimize and usually do not incorporate the known structure of the target distribution
due to their black-box nature.
To expand the space of flexible variational families, 
we revisit Variational Rejection Sampling (VRS) \citep{grover2018variational}, which combines a parametric
proposal distribution with rejection sampling to define a rich non-parametric family of distributions that
explicitly utilizes the known target distribution.
By introducing a low-variance reparameterized gradient estimator for the parameters of the proposal distribution,
we make VRS an attractive inference strategy for models with continuous latent variables. 
We argue theoretically and demonstrate empirically that the resulting method---Reparameterized Variational 
Rejection Sampling (RVRS)---offers an attractive trade-off between computational cost and inference fidelity.
In experiments we show that our method performs well in practice and that it is
well-suited for black-box inference, especially for models with local latent variables. % in probabilistic programming frameworks.
\end{abstract}

%%%%%%%%%%%%%%%%%%%%%%%%%%%%%%%%%%%%%
\section{Introduction}
%%%%%%%%%%%%%%%%%%%%%%%%%%%%%%%%%%%%%

Variational inference is a powerful method for approximate Bayesian inference with a number of appealing
properties, including support for data subsampling and model learning \citep{blei2017variational}. 
Unfortunately, simple variational families like mean-field gaussian distributions often
result in poor posterior approximations, while defining custom parametric families that better reflect
the correlation structure and tail behavior of the exact posterior can be difficult, even for experts.
This has motivated research into more flexible variational methods, including black-box methods like
normalizing flows \citep{rezende2015variational} as well as hybrid methods that incorporate Markov Chain Monte Carlo (MCMC) \citep{salimans2015markov}.

While these methods are powerful, they come with several disadvantages. Normalizing flows can be difficult
to optimize, exhibit tail behavior that is difficult to control \citep{jaini2020tails}, 
and introduce a large design space characterized by many hard-to-set hyperparameters.
Moreover, due to their black-box nature normalizing flows typically
do not incorporate the known structure of the target distribution. This is arguably a lost opportunity, especially
in the context of probabilistic programming systems, where this information is readily available.
The most powerful methods that combine variational inference with MCMC are gradient-based
\citep{geffner2021mcmc,zhang2021differentiable,thin2021monte}, with the
result that many (possibly expensive) gradient steps may be required to generate a single sample. Moreover,
good performance relies on carefully tuning the MCMC kernel, which can be challenging, since
posterior curvature can vary considerably across latent space. In addition,
these approaches typically introduce auxiliary latent variables, leading to a looser
and more stochastic variational bound.

These considerations lead us to revisit a conceptually simpler hybrid variational inference method dubbed
Variational Rejection Sampling (VRS) \citep{grover2018variational}. Like MCMC-based methods,
the target distribution is directly incorporated into the definition of the variational family, resulting
in a non-parametric variational distribution. Since, however, rejection sampling is much simpler than MCMC, 
the result is a considerably simpler hybrid variational method that does not require delicate tuning or
differentiating through long MCMC chains.
Unfortunately, VRS utilizes score function (i.e.~REINFORCE-like \citep{williams1992simple}) gradient estimators,
which are known to be high variance, thus limiting its usefulness to discrete latent variable models,
which are in any case not amenable to the reparameterization trick.
In this work we set out to show that by introducing a reparameterized gradient estimator VRS 
becomes an attractive inference strategy for continuous latent variable models.

In summary our contributions include the following:
%%%%
\begin{enumerate}[itemsep=0.5pt]
\item We introduce a reparameterized gradient estimator for VRS.
\item We show that the resulting method---RVRS---is especially well-suited for local latent variable models,
    including hierarchical models that additionally include global latent variables.
\item We characterize the variational gap of (R)VRS as a function of the rejection threshold parameter $T$.
%\item We provide an open source implementation of our methods at \texttt{https://github.com/anon}. 
\end{enumerate}
%%%%

%%%%%%%%%%%%%%%%%%%%%%%%%%%%%%%%%%%%%
\section{Problem setting}
\label{sec:setting}
%%%%%%%%%%%%%%%%%%%%%%%%%%%%%%%%%%%%%

We are given a model with joint density of the form $\pth(\bx, \bz) = \pth(\bx | \bz) \pth(\bz)$
where the latent variable $\bz \in \RR^D$ is governed
by a prior $\pth(\bz)$ and $\bx$ in the likelihood $\pth(\bx | \bz)$ 
represents observed data.
We aim to devise a flexible variational approximation to the
posterior $\pth(\bz | \bx)$ that can be learned with a low-variance ELBO gradient estimator. 
Initially we do not assume any particular conditional independence structure, but in \secref{sec:hier}
we turn our attention to hierarchical models with both global and local latent variables, which benefit from additional
consideration.
We would like our method to be generic in nature so that it is suitable for black-box inference
in a probabilistic programming framework.
Additionally we would like our method to support model learning, 
i.e.~learning $\bth$ in conjunction with the approximate posterior.

%%%%%%%%%%%%%%%%%%%%%%%%%%%%%%%%%%%%%
\section{Background}
\label{sec:bg}
%%%%%%%%%%%%%%%%%%%%%%%%%%%%%%%%%%%%%

%%%%%
\subsection{Variational inference}
%%%%%

The most common variant of variational inference introduces a parametric
variational distribution $\qphi(\bz)$ and proceeds to optimize the parameters $\bphi$
to minimize the Kullback-Leibler (KL) divergence between $\qphi(\bz)$ and the posterior $\pth(\bz | \bx)$,
i.e.~$\KL( \qphi(\bz) || \pth(\bz | \bx))$.
This can be done by maximizing the Evidence Lower Bound or ELBO
%%%
\begin{equation}
    \label{eqn:elbo}
    \elbo \equiv \EE_{\qphi(\bz)} \left[ \log \pth(\bx, \bz) - \log \qphi(\bz)\right]
    \le \log \pth(\bx) \equiv \log \EE_{\pth(\bz)} \left [ \pth(\bx|\bz) \right ]
\end{equation}
%%%
Thanks to the inequality in Eqn.~\ref{eqn:elbo} the ELBO naturally enables joint model learning and inference, i.e.~we can
maximize the ELBO w.r.t.~both variational parameters $\bphi$ and model parameters $\bth$ simultaneously.
As noted in the introduction, a potential shortcoming of this fully parametric approach is the difficulty of 
specifying suitable parameterizations for $\qphi(\bz)$. For additional background see e.g.~\citep{blei2017variational}.

%%%%%
\subsection{Variational Rejection Sampling}
%%%%%
The basic idea behind VRS is simple: define a flexible variational distribution by taking a parametric
proposal distribution $\qphi(\bz)$ and warping it towards the posterior $\pth(\bz | \bx)$ via a smoothed
variant of rejection sampling. In more detail, define the variational distribution $\rr(\bz)$ as
%%%
\begin{equation}
\label{eqn:rdef}
\rr(\bz) \equiv \frac{\qphi(\bz) \aaa(\bz)}{\ZZr} \qquad
    {\rm with} \qquad 
\ZZr \equiv \int \! d \bz \; \qphi(\bz) \aaa(\bz)
\end{equation}
%%%
where
%%%
\begin{equation}
\aaa(\bz) \equiv \sigma(\log \pth(\bx, \bz) - \log \qphi(\bz) + T) = \sigma(-\ellT(\bz) )
\end{equation}
%%%
is an acceptance probability with $\aaa(\bz) \in [0, 1]$.
Here $\sigma(\cdot)$ is the logistic function and $T \in \RR$ is a threshold parameter.
Moreover we have defined the $T$-shifted log ratio
%%%
\begin{equation}
\ellT(\bz) \equiv -\log \pth(\bx, \bz) + \log \aaa(\bz) - T
\end{equation}
%%%
As $T\rightarrow \infty$ we have $\aaa(\bz) \rightarrow 1$ and $\rr(\bz) \rightarrow \qphi(\bz)$,
recovering conventional variational inference with $\qphi(\bz)$ as the variational distribution.
In the opposite limit $T\rightarrow -\infty$ the acceptance probability is low, $\aaa(\bz) \rightarrow 0$,
and $\rr(\bz) \rightarrow \pth(\bz | \bx)$.
For intermediate $T$ (i.e.~$T$ which leads to a few but not many rejected samples) we get
a $\rr(\bz)$ that is closer to the posterior $\pth(\bz | \bx)$ than the proposal distribution $\qphi(\bz)$ at the cost of a moderate amount of additional computation. Indeed as shown in \citet{grover2018variational}, as $T$ decreases
for fixed $\qphi(\bz)$ the ELBO increases monotonically and thus the Kullback-Leibler divergence 
$\KL(\rr(\bz) || \pth(\bz | \bx))$ decreases monotonically.

%%%%%%%%%%%%%
\subsubsection{Sampling}
%%%%%%%%%%%%%
Since $\aaa(\bz) \in [0, 1]$ it is straightforward to sample from $\rr(\bz)$, see Algorithm~\ref{alg:sample}.
The expected number of draws from the proposal distribution is given by $\ZZr^{-1}$, see \secref{sec:samplingcost}.
For this reason we expect the sweet spot for VRS to occur for moderate values of $\ZZr^{-1} \sim 3-10$, where the cost of 
rejection sampling is not too high but where the proposal distribution is still significantly `sculpted' towards
the posterior.

\begin{algorithm}[t]
\caption{Sampler for $\rr(\bz)$.
    {\bf Input}: acceptance~probability~$\aaa(\bz)$ and proposal $\qphi(\bz)$.
    }
\label{alg:sample}
\begin{algorithmic}[1]
\While{True}
\State $\bz \sim \qphi(\bz)$
    \If {$u < \aaa(\bz)$ where $u \sim {\rm Uniform}(0,1)$}
\State return $\bz$
\EndIf
\EndWhile
\end{algorithmic}
\end{algorithm}

%%%%%%%%%%%%%
\subsubsection{Gradient estimators}
\label{sec:vrsgradest}
%%%%%%%%%%%%%

VRS is only practical if we can use gradient methods to optimize the corresponding ELBO given by
%%%
\begin{align}
\label{eqn:rvrselbo}
\elbo(\bphi, \bth) = \EE_{r_{\bth, \bphi}(\bz)} \left[ \log \pth(\bx, \bz) - \log \rr(\bz) \right]
\end{align}
%%%
As shown in \citet{grover2018variational}, gradients for the parameters $\bphi$ that define the proposal 
distribution $q_{\bphi}(\bz)$ can be computed using the following estimator
%%%
\begin{align}
\label{eqn:phigrad}
\nabla_\bphi \elbo = {\rm COV}_{\rr(\bz)} \left[ \AAA(\bz), \aaa(\bz) \nabla_\bphi \log \qphi(\bz) \right]
\end{align}
%%%
with
%%%
\begin{equation}
\AAA(\bz) \equiv \log \pth(\bx, \bz) - \log \qphi(\bz) - \log \aaa(\bz)
\end{equation}
%%%
and where ${\rm COV}_{\rr(\bz)}[A(\bz), B(\bz)]$ denotes the covariance between random variables $A$ and $B$
w.r.t.~the distribution $\rr(\bz)$.
%%%%%%%%%%%%
Similarly the gradient estimator for the model parameters $\bth$ is given by
%%%
\begin{align}
\label{eqn:thetagrad}
\nabla_\bth \elbo = \EE_{\rr(\bz)} \left[ \nabla_\bth \log \pth(\bx, \bz) \right] -
 {\rm COV}_{\rr(\bz)} \left[\AAA(\bz), (1 - \aaa(\bz)) \nabla_\bth \log \pth(\bx, \bz) \right]
\end{align}
%%%
It is easy to show (see \secref{sec:vrsgradestsupp}) that in the limit that $\aaa(\bz) \rightarrow 1$ and $\rr(\bz) \rightarrow \qphi(\bz)$
the gradient estimator \eqref{eqn:phigrad} reduces to a conventional score function
(i.e.~REINFORCE-like) gradient estimator, which is known to exhibit high variance, essentially due to its 
coarse credit assignment \citep{mohamed2020monte}. It is straightforward to compute unbiased Monte Carlo estimates of \eqref{eqn:phigrad} and \eqref{eqn:thetagrad}, although doing so requires drawing $S > 1$ samples from $\rr(\bz)$ due to the covariance terms, see \secref{sec:vrsmc}.

%%%%%%%%%%%%%%%%%%%%%%%%%%%%%%%%%%%%%
\section{Reparameterized Variational Rejection Sampling}
\label{sec:rvrs}
%%%%%%%%%%%%%%%%%%%%%%%%%%%%%%%%%%%%%

The REINFORCE-like covariance term in \eqnref{eqn:phigrad} %and \eqnref{eqn:thetagrad}
is generally expected to be high variance
and thus limit the applicability of VRS.
Fortunately, as we show in Prop.~\ref{prop:gradest}, the VRS ELBO admits a reparameterized (i.e.~pathwise) gradient estimator for $\bphi$ if $\qphi(\bz)$ is reparameterizable---a surprising capability, since $\rr(\bz)$ is not readily reparameterizable itself.
Since the suite of reparameterizable proposal distributions is quite large---including e.g.~Normal distributions, 
Dirichlet distributions, and normalizing flows with reparameterizable base distributions---the RVRS distribution
$\rr(\bz)$ is quite flexible.
%%%
\begin{restatable}{prop}{gradest}
If the proposal distribution $\qphi(\bz)$ is reparameterizable, then
the VRS ELBO \eqnref{eqn:rvrselbo} admits the following reparameterized gradient estimator for $\bphi$ gradients
%%%
\begin{align}
\label{eqn:repphigrad}
\nabla_\bphi \elbo = \EE_{\rr(\bz)} \left[\left( 2 \AAAbar(\bz) \frac{\partial \aaa(\bz) }{\partial \bz} + \aaa(\bz) \frac{\partial \AAA(\bz)  }{\partial \bz}
 \right) \cdot \nabla_{\bphi} \bz \right]
\end{align}
%%%
where $\AAAbar(\bz)$ is defined as
$\AAAbar(\bz) \equiv  \AAA(\bz) - \EE_{\rr(\bz^\prime)}\left[ \AAA(\bz^\prime)\right]$
    and $\nabla_{\bphi} \bz$ is the velocity field\footnote{For example if $\qphi(z) = \NN(z | \mu, \sigma^2)$ 
    then $\nabla_\mu z = 1$ and $\nabla_\sigma z = (z - \mu)/\sigma $.} corresponding to infinitesimal 
    displacement of $\qphi(\bz)$ in $\bphi$-space.
\label{prop:gradest}
\eqnref{eqn:repphigrad} reduces to a conventional reparameterized gradient
in the limit that $\aaa(\bz) \rightarrow 1$ and $\rr(\bz) \rightarrow \qphi(\bz)$.
See \secref{sec:phiest} for the proof and additional details.\footnote{In particular in \secref{sec:rvrsautogradsupp} 
we describe how we leverage automatic differentation and $S>1$ samples from $\rr(\bz)$ to obtain an unbiased 
    Monte Carlo estimate of \eqnref{eqn:repphigrad}.}
\end{restatable}
%%%
Fundamentally the existence of a pathwise gradient estimator can be traced to three properties
of $\rr(\bz)$: i) $\rr(\bz)$ is proportional to a reparameterizable distribution, namely $\qphi(\bz)$;
ii) $\rr(\bz)$ depends on $\bphi$ only through $\qphi(\bz)$; and iii) we can compute $\aaa(\bz)$ and its gradients pointwise.
We note that the derivation of Prop.~\ref{prop:gradest} is conceptually similar to that behind 
`doubly reparameterized gradients' \citep{tucker2018doubly}, although in that case a gradient estimator
that is \emph{already} reparameterized is manipulated to transform a score-function-like term to further reduce variance.

%%%%%%%%%%%%%%%%%%%%%%%%%%%%%%%%%%%%
\subsection{Model parameter gradients}
\label{sec:thetagraddisc}
%%%%%%%%%%%%%%%%%%%%%%%%%%%%%%%%%%%%

Unfortunately it seems unlikely that the covariance term in \eqnref{eqn:thetagrad} can be 
reparameterized in a straightforward way, since eliminating $\nabla_\bth \log \pth(\bx, \bz)$ would
require e.g.~a reparameterized sampler of $\pth(\bz | \bx)$. 
However, we show empirically that this term can be safely dropped at the cost of introducing some bias. 
This is because this term encodes how the log evidence estimate changes
due to changes in $\ellT(\bz)$ and not the `direct' change encoded by the term $\EE_{\rr(\bz)} \left[ \nabla_\bth \log \pth(\bx, \bz) \right]$.

%%%%%%%%%%%%%%%%%%%%%%%%%%%%%%%%%%%%
\subsection{Adapting the threshold $T$}
\label{sec:tuning}
%%%%%%%%%%%%%%%%%%%%%%%%%%%%%%%%%%%%

Choosing an appropriate value of $T$ in the vicinity of 
$\EE_{\qphi(\bz)} \left[ \log \qphi(\bz) - \log \pth(\bx, \bz)\right]$
is crucial for good performance of (R)VRS. In \citet{grover2018variational} the authors
propose a strategy based on quantiles of $\log \pth(\bx, \bz) / \qphi(\bz)$.
While we find that this strategy can work, we prefer
a gradient-based strategy for tuning the threshold parameter $T$ that allows direct control over
the computational cost of (R)VRS. Another advantage of this approach is that because it is gradient-based
it offers the possibility of choosing $T$ using amortized inference, although we do not explore that possibility here.
Recall that $\ZZr \equiv \int \! d \bz \; \qphi(\bz) \aaa(\bz)$ is the mean acceptance probability of the rejection sampler
and consider the loss $\LL(T) = \tfrac{1}{2} \left(\ZZr - \ZZ_\target \right)^2$
where $\ZZ_\target \in (0, 1)$ is a target acceptance probability. 
Then the gradient $\frac{\partial \LL}{\partial T}$ is given by
%%%
\begin{align}
\frac{\partial \LL}{\partial T} \!=\! \left(\ZZ_r \! - \! \ZZ_\target \right) \! \EE_{\qphi(\bz)}\!\left[ \frac{\partial \aaa(\bz)} {\partial T} \right] 
\!= \EE_{\qphi(\bz)}\left[ \aaa(\bz) \! - \! \ZZ_\target \right] \EE_{\qphi(\bz)}\left[ \aaa(\bz) (1 \!-\! \aaa(\bz))\right]
 \end{align}
%%%
which we can readily compute unbiased estimates of, since we have $S>1$ samples at our disposal.
Throughout this work we use MC estimates of $\frac{\partial \LL}{\partial T}$ to tune $T$; 
see \secref{sec:apptuning} in the supplement for details.

%%%%%%%%%%%%%%%%%%%%%%%%%%%%%%%%%%%%
\subsection{Models with only local latent variables}
\label{sec:localmodels}
%%%%%%%%%%%%%%%%%%%%%%%%%%%%%%%%%%%%

For models with only local latent variables like a VAE \citep{kingma2013auto} sampling, ELBO estimation,
and ELBO gradient estimation for RVRS trivially factorize across data points, and thus RVRS admits unbiased mini-batch learning for such models.
An efficient sampler for RVRS in this scenario requires a flexible rejection sampling scheme
that maximizes usage of computational resources. In particular during training we can choose between: 
i) an unbiased sampler that terminates when $S>1$ latent samples have been generated for every data point; and 
ii) a (potentially much) faster biased sampler that terminates after generating a \emph{fixed}
number of proposals for each data point.
See Algorithm~\ref{alg:semiunbiased} \& Algorithm~\ref{alg:semibiased} in the supplement for details. 
As we report in \figref{fig:biasedsemi} in \secref{sec:addexp} the small bias introduced by the faster sampler has
a correspondingly small impact on performance.

%%%%%%%%%%%%%%%%%%%%%%%%%%%%%%%%%%%%
\subsection{Hierarchical models with global and local latent variables}
\label{sec:hier}
%%%%%%%%%%%%%%%%%%%%%%%%%%%%%%%%%%%%

We now consider models with both a global latent variable $\bz_G$ and local latent variables $\{ \bz_n \}$, 
with $n=1,...,N$ indexing the $N$ observed data points $\{ \bx_n \}$. 
We assume the following conditional independence structure:
%%%
\begin{align}
    \pth(\bx_{1:N}, \bz_G, \bz_{1:N}) = \pth(\bz_G) \prod_{n=1}^N \pth(\bx_n | \bz_n, \bz_G) p(\bz_n | \bz_G)
\end{align}
%%%
While RVRS can be applied to the joint latent space $\{\bz_G, \bz_{1:N}\}$, the resulting algorithm
does not admit unbiased data subsampling (i.e.~mini-batch learning), since $\aaa(\bz)$ %RVRS acceptance probability
depends on the entire dataset, limiting this approach to moderate $N$.\footnote{This is
of course equally true of other non-parametric approaches like UHA/DAIS \citep{geffner2021mcmc,zhang2021differentiable},
although see \citep{jankowiak2022surrogate}.}
To enable data subsampling we adopt a hybrid approach in which the posterior
over $\bz_G$ is approximated by a parametric distribution $\qphi(\bz_G)$ while the conditional
posteriors $\pth(\bz_n | \bz_G, \bx_n)$ are approximated by RVRS. This can be understood
as an instance of a `locally enhanced bound' \citep{geffner2022variational}, and
is analogous to the `Semi-DAIS' approach explored in \citet{jankowiak2022surrogate} in the context of UHA/DAIS. 
We refer to this semi-parametric approach as Semi-RVRS. See \secref{sec:semisupp} for details.

%%%%%%%%%%%%%%%%%%%%%%%%%%%%%%%%%%%%%
\section{Convergence analysis}
\label{sec:theory}
%%%%%%%%%%%%%%%%%%%%%%%%%%%%%%%%%%%%%

It is evident from the structure of $\rr(\bz)$ in \eqnref{eqn:rdef} that as $T \rightarrow -\infty$ the 
variational distribution $\rr(\bz)$ converges to the exact posterior $\pth(\bz | \bx)$ \emph{pointwise}. 
But can we say anything about the corresponding ELBO in \eqnref{eqn:rvrselbo}? 
As we would expect, the variational gap goes to zero in the same limit as $e^T$, see Prop.~\ref{prop:vargap}.
Notably the relative simplicity of rejection sampling allows us to prove a generic result, whereas an analogous result
for DAIS in \citep{zhang2021differentiable} is limited to linear Gaussian models due to the complexity of analyzing MCMC chains.
%%%
\begin{restatable}{prop}{vargap}
\label{prop:vargap}
%Assume that $p_\bth(\bx, \bz)$ and $q_\bphi(\bz)$ are positive for all $\bz$ and that
{\bf (A)} 
Assume that
 $q_{\bphi}(\bz)$ is sufficiently heavy-tailed so that $\xi \equiv \EE_{p_\bth(\bz|\bx)} \left[  \frac{ p_\bth(\bx,\bz)}{q_{\bphi}(\bz)}  \right]$ is finite.
Then the variational gap $\Delta$ between $ \log  p_\bth(\bx)$
and the ELBO is bounded from above as $\Delta < \frac{3}{2} e^T \xi$ for $T < -\log 2 \xi$.
{\bf (B)} 
    An analogous bound holds for the hierarchical modeling case considered in \secref{sec:hier}, where the bound
includes an additional term $\KL \left( \qphi(\bz_G) \Big | \Big | \; \pth(\bz_G | \bx_{1:N}) \right)$ that encodes
the suboptimality of the parametric variational approximation for the global latent variable $\bz_G$.
For additional details and the proof see \secref{sec:globalproof} in the supplement.
\end{restatable}
%%%

%%%%%%%%%%%%%%%%%%%%
\begin{figure}[ht] \begin{center}
\includegraphics[width=0.99\textwidth]{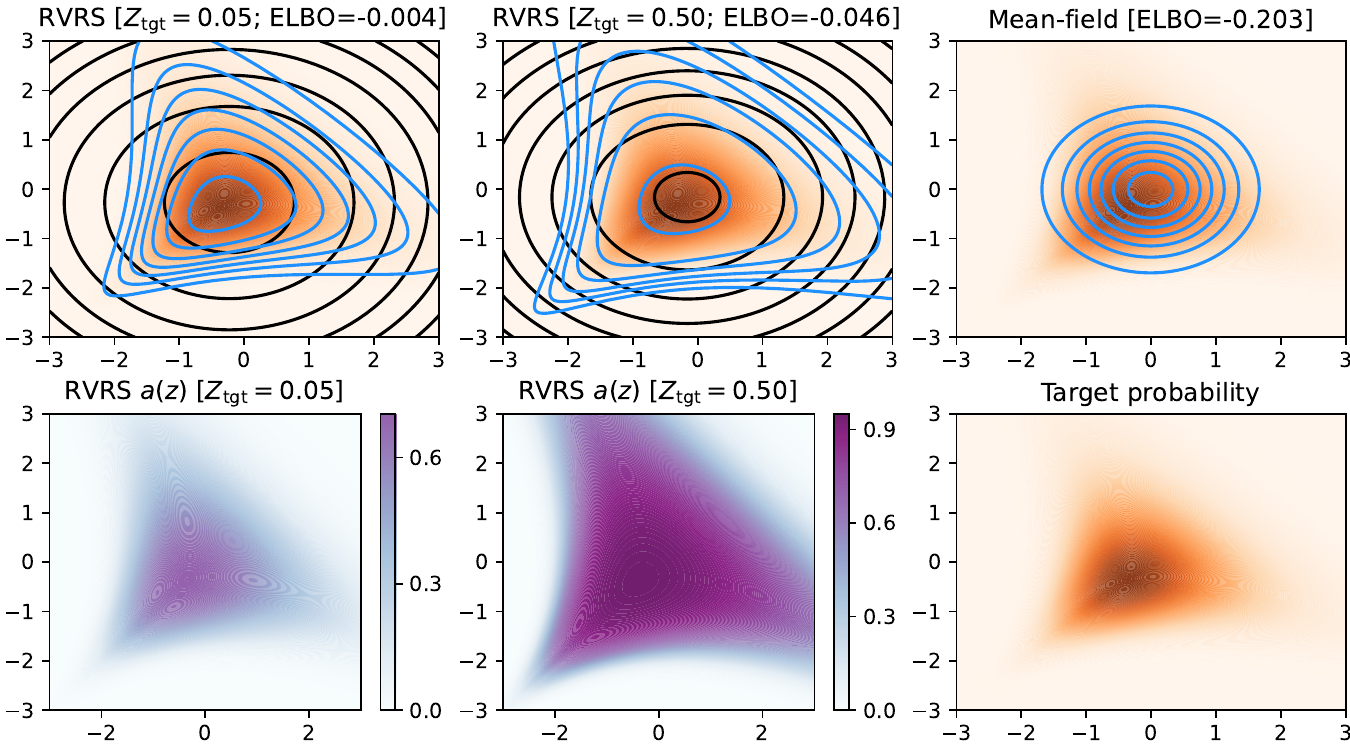}
    \caption{
    We illustrate how RVRS works on a (normalized) funnel-shaped target distribution (orange density).
    Blue contours depict variational fits, with a mean-field Normal fit depicted in the upper right figure.
    The first two columns depict RVRS fits for $\ZZ_\target \in \{0.05,0.5\}$,
    with black contours depicting mean-field Normal proposal distributions $\qphi(\bz)$.
    The leftmost figures in the lower row depict the acceptance probability $\aaa(\bz)$;
    for $\ZZ_\target=0.05$ $\aaa(\bz)$ differs significantly from $1$ everywhere so that
    $\rr(\bz)$ is strongly sculpted towards the target and the ELBO is nearly optimal (i.e.~close to $0$).
    } \label{fig:funnel}
\end{center} \end{figure}
%%%%%%%%%%%%%%%%%%%%

%%%%%%%%%%%%%%%%%%%%%%%%%%%%%%%%%%%%%
\section{Related Work}
\label{sec:related}
%%%%%%%%%%%%%%%%%%%%%%%%%%%%%%%%%%%%%

Many variational objectives that go beyond a conventional ELBO have been proposed in the literature.
These include the importance weighted autoencoder (IWAE) \citep{burda2015importance,cremer2017reinterpreting},
the thermodynamic variational objective \citep{masrani2019thermodynamic},
and approaches that make use of Sequential Monte Carlo \citep{le2017auto,maddison2017filtering,naesseth2018variational}.
Variational Rejection Sampling (VRS) was proposed by \citet{grover2018variational} and applied to models
with discrete latent variables.
An early combination of MCMC methods with variational inference was proposed by \citet{salimans2015markov} and
\citet{wolf2016variational} and has led to follow-up work by many authors \citep{hoffman2017learning,caterini2018hamiltonian,ruiz2019contrastive}. Arguably the most powerful hybrid variational methods proposed so far are those that incorporate 
gradient-based MCMC like Uncorrected Hamiltonian Annealing (UHA; \citep{geffner2021mcmc}) and the essentially identical algorithm
Differentiable Annealed Importance Sampling (DAIS; \citep{zhang2021differentiable}); see also \citep{thin2021monte,doucet2022score,pmlr-v162-matthews22a}. A conceptually related but distinct gradient-based approach utilizes ergodic maps built with Hamiltonian dynamics to formulate
flexible variational distributions \citep{xu2023mixflows,xu2023embracing}.
For a recent review of some of these methods see \citet{doucet2023differentiable}.
Another important line of work has seen the development of rich parametric families of distributions like normalizing flows for use in 
variational inference \citep{rezende2015variational,kingma2016improved,papamakarios2021normalizing}.
Rejection sampling has seen other applications in probabilistic machine learning. For example \citet{stimper2022resampling} adapt earlier work \citep{bauer2019resampled} to build normalizing flows where the base distribution is defined via a
 learned rejection sampling scheme. Indeed \citet{stimper2022resampling} use a REINFORCE-like gradient estimator
 modified from VRS that could benefit from our reparameterized estimator in Prop.~\ref{prop:gradest}. 
Finally \citet{naesseth2017reparameterization} show how to construct partially reparameterized gradient
estimators for distributions defined by classical rejection samplers (i.e.~not the `smoothed' variant in \eqnref{eqn:rdef}).

%%%%%%%%%%%%%%%%%%%%%%%%%%%%%%%%%%%%%
\section{Experiments}
\label{sec:exp}
%%%%%%%%%%%%%%%%%%%%%%%%%%%%%%%%%%%%%

All our experiments are implemented using JAX and NumPyro \citep{bradbury2020jax,phan2019composable,bingham2019pyro}.
We explore a number of different aspects of RVRS, 
including support for $D\gg1$ latent dimensions and model learning (\secref{sec:gp}), 
variational auto-encoders (\secref{sec:vae}), and
hierarchical models (\secref{sec:local}). 
We provide additional experimental details and report additional results in \secref{sec:expdetails}-\ref{sec:addexp}.

%%%%%%%%%%%%%%%%%%%%%%%%%%%%%%%%%%%%%%%%%%%%%%%%%%
\subsection{Characterizing RVRS}
\label{sec:char}
%%%%%%%%%%%%%%%%%%%%%%%%%%%%%%%%%%%%%%%%%%%%%%%%%%

We begin with a few experiments to characterize some of the general characteristics of RVRS.
In \figref{fig:funnel} we illustrate graphically how RVRS `sculpts' a mean-field gaussian proposal distribution $\qphi(\bz)$ to
match a non-gaussian target. Notably a nearly optimal ELBO is achieved for $\ZZ_\target=0.05$.

\begin{figure}[ht] \begin{center}
\includegraphics[width=0.4\textwidth]{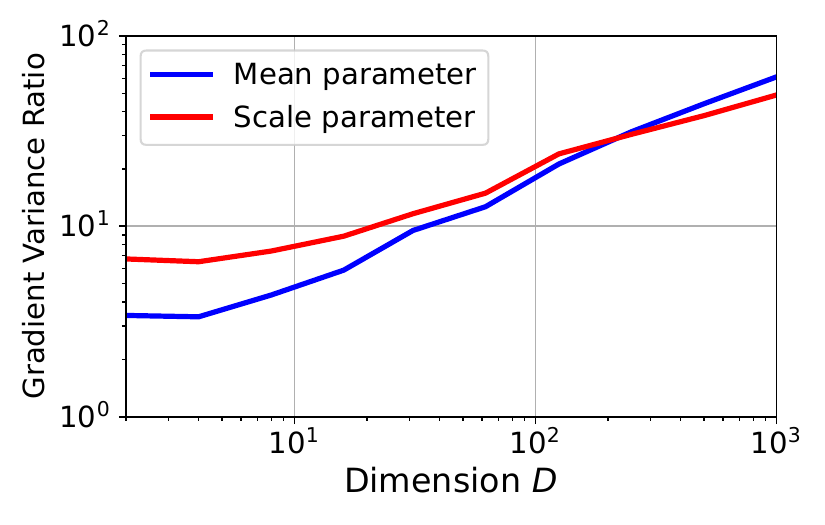}
    \caption{ 
    We compare RVRS and VRS gradient variance for a logistic regression problem with $N=100$ data points
    as we vary the latent dimension $D$. The proposal $\qphi(\bz)$ is mean-field Normal and we depict
    the ratio of gradient variances between VRS and RVRS for the mean and scale (i.e.~root variance) parameters of $\qphi(\bz)$.
    } \label{fig:gradvar}
\end{center} \end{figure}

Next we compare the variance of RVRS and VRS ELBO gradient estimators on a logistic regression model, see \figref{fig:gradvar}.
We find that VRS gradient variance is always larger than in the case of RVRS---e.g.~by a factor of $\sim15$ for
$D=62$ latent dimensions---and that the ratio increases as the dimension increases. For an example
of how large gradient variance negatively impacts the optimization performance of VRS see \figref{fig:elbocurves} in \secref{sec:addexp}.

\begin{figure}[ht]
\begin{center}
\includegraphics[width=0.5\textwidth]{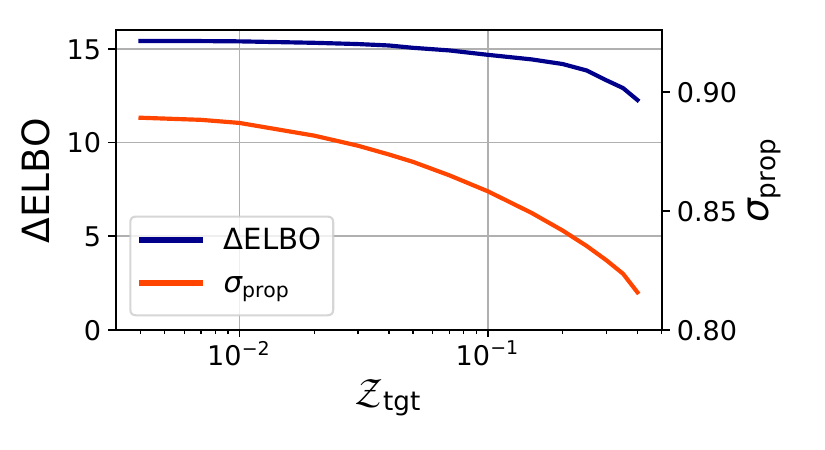}
    \caption{ 
    We explore the performance of RVRS as a function of $\ZZ_\target$ on a logistic
    regression problem in $D=51$ dimensions. The blue curve depicts the ELBO improvement over
    a mean-field baseline, while the orange curve depicts the geometric mean of the $D$ scales (i.e.~root variances)
    that define the mean-field Normal proposal $\qphi(\bz)$. As $\ZZ_\target \rightarrow 0$ %and $T\rightarrow -\infty$
    the proposal distribution becomes broader, especially compared to the mean-field fit 
    obtained with a standard ELBO, which yields $\sigma = 0.70$.
    }
    \label{fig:varyZ}
\end{center}
\end{figure}

Finally in \figref{fig:varyZ} we explore how RVRS depends on the hyperparameter $\ZZ_\target \in (0, 1)$.
As we would expect the ELBO increases monotonically as $\ZZ_\target$ decreases---as it must
if $T$ adaptation and $\qphi(\bz)$ learning are working correctly. Tellingly, we
see that the width of the proposal distribution $\qphi(\bz)$ increases as $\ZZ_\target$ decreases.
This illustrates the basic principle that (R)VRS exploits to achieve better variational approximations.
Since the ELBO tends to prefer variational distributions that excessively avoid low-density regions of the posterior,
a common failure mode of parametric variational inference is to underestimate posterior uncertainty.
Target-dependent rejection sampling offers a simple but effective mechanism to better capture posterior uncertainty: inflate
the width of the proposal distribution where needed and reject a portion of samples in regions where the density of the proposal
$\qphi(\bz)$ is excessive (due to e.g.~the parametric misfit of the proposal). 
The upshot is that (R)VRS can better capture tail behavior and thus yield higher
fidelity variational approximations.

%%%%%%%%%%%%%%%%%%%%
%%%%%%%%%%%%%%%%%%%%

%%%%%%%%%%%%%%%%%%%%
%%%%%%%%%%%%%%%%%%%%

%%%%%%%%%%%%%%%%%%%%%%%%%%%%%%%%%%%%%%%%%%%%%%%%%%
\subsection{Logistic regression}
\label{sec:logreg}
%%%%%%%%%%%%%%%%%%%%%%%%%%%%%%%%%%%%%%%%%%%%%%%%%%

\begin{figure}[ht]
\begin{center}
\includegraphics[width=1.04\textwidth]{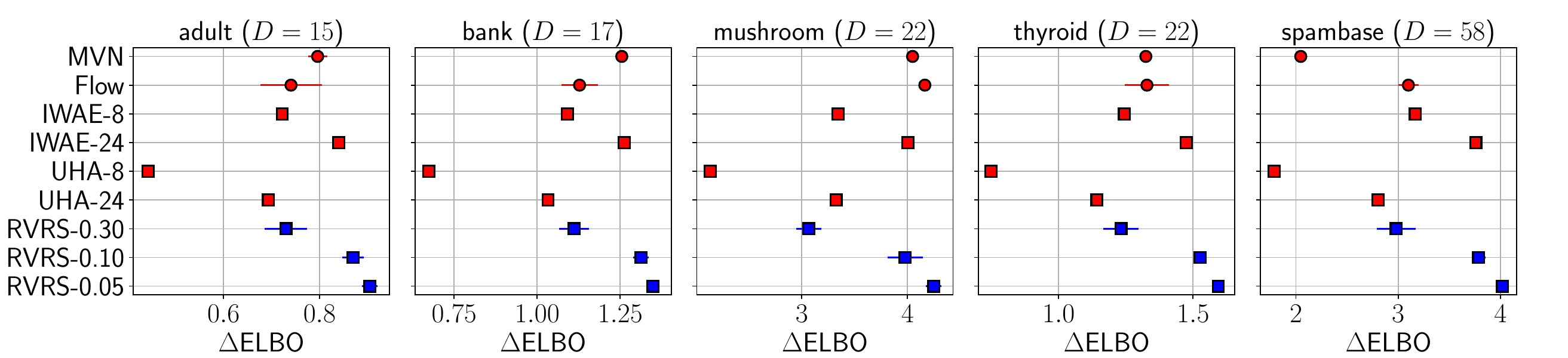}
\includegraphics[width=0.50\textwidth]{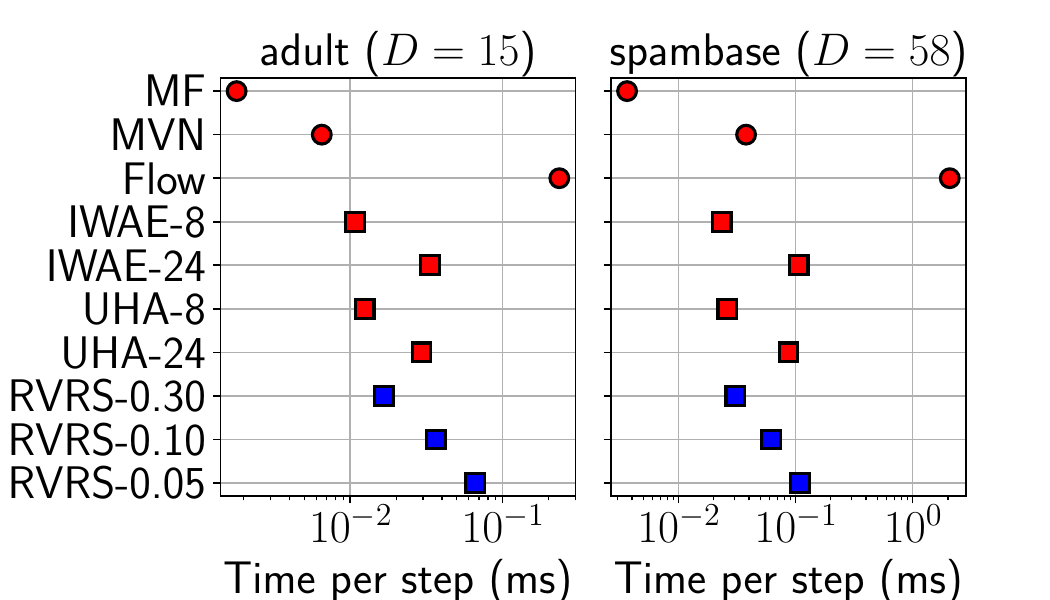}
\includegraphics[width=0.48\textwidth]{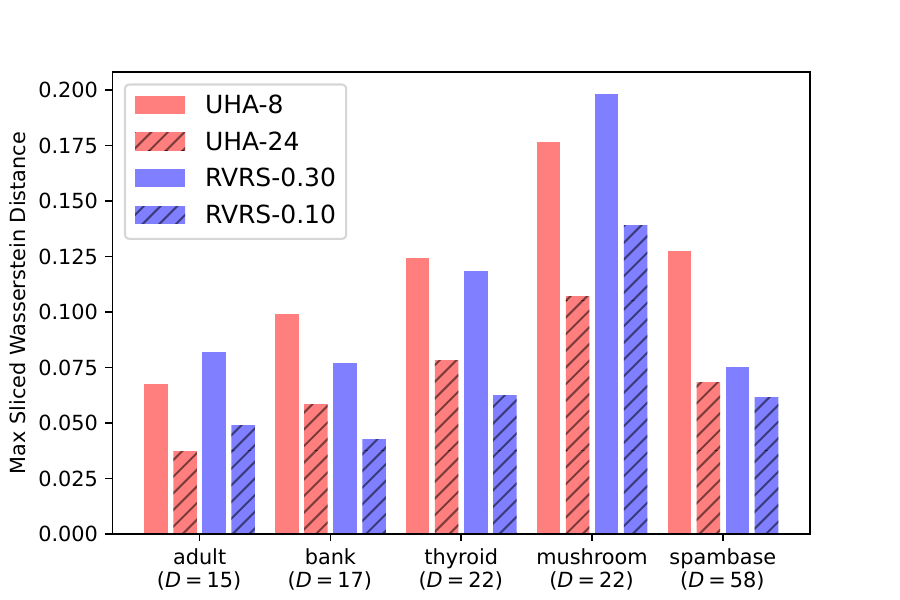}
    \caption{ 
    {\bf (Top)} We depict ELBO improvements above the mean-field baseline for $9$ variational methods on $5$
    logistic regression tasks. Circles and squares indicate parametric and non-parametric methods, respectively. Error bars denote two standard deviations and ELBOs are averaged across $5$ runs.
    {\bf (Bottom left)} We depict the corresponding gradient step times for two logistic regression tasks. 
    {\bf (Bottom right)} We compare the fidelity of posterior samples generated by UHA and RVRS w.r.t.~the Max Sliced Wasserstein distance,
    using samples from NUTS as a reference. Here and elsewhere RVRS-$0.50$ refers to RVRS with $\ZZ_\target=0.5$,
    IWAE-$8$ refers to IWAE with $K=8$ particles, etc.
    }
    \label{fig:logistic}
\end{center}
\end{figure}

We compare RVRS to a large number of variational baselines on $5$ logistic regression tasks. To ensure
that posterior distributions are relatively non-gaussian we consider $N=100$ data points, while the latent
dimension ranges from $D=15$ to $D=58$. 
We consider three fully parametric baselines: 
mean-field with a factorized Normal distribution ({\bf MF});
a multivariate Normal distribution ({\bf MVN}); 
and a Block Neural Autoregressive normalizing flow ({\bf Flow}; \cite{de2020block}). 
We also consider four non-parametric baselines: 
IWAE with $K \in \{8,24\}$ particles ({\bf IWAE}); and
UHA with $K \in \{8,24\}$ gradient steps ({\bf UHA}). For RVRS we consider $\ZZ_\target \in \{0.3, 0.1, 0.05\}$.
For the results see \figref{fig:logistic}.

We find that RVRS performs well across the board. For example RVRS with $\ZZ_\target = 0.10$ outperforms
the normalizing flow on $4$/$5$ datasets but is much faster to train. Moreover RVRS-$0.10$ matches
or exceeds the performance of IWAE with $K=24$ particles on all datasets. The RVRS-$0.10$ ELBO also exceeds
that of UHA with $K=24$ steps on all datasets, but we note that this gap is probably at least partially explained
by the additional looseness of the UHA variational bound. Indeed if we compare RVRS and UHA posterior
samples to `gold standard' samples obtained with NUTS \citep{hoffman2014no,carpenter2017stan} and use a Max Sliced
Wasserstein distance \citep{deshpande2019max} to quantify the fidelity of the posterior approximation,
we find that RVRS-$0.10$ (respectively,~RVRS-$0.30$) approximately matches the performance of UHA-$24$ (respectively,~UHA-$8$),
see \figref{fig:logistic}.

%%%%%%%%%%%%%%%%%%%%
%%%%%%%%%%%%%%%%%%%%

%%%%%%%%%%%%%%%%%%%%%%%%%%%%%%%%%%%%%%%%%%%%%%%%%%
\subsection{Gaussian process classification}
\label{sec:gp}
%%%%%%%%%%%%%%%%%%%%%%%%%%%%%%%%%%%%%%%%%%%%%%%%%%

\begin{figure}[ht]
\begin{center}
\includegraphics[width=0.8\textwidth]{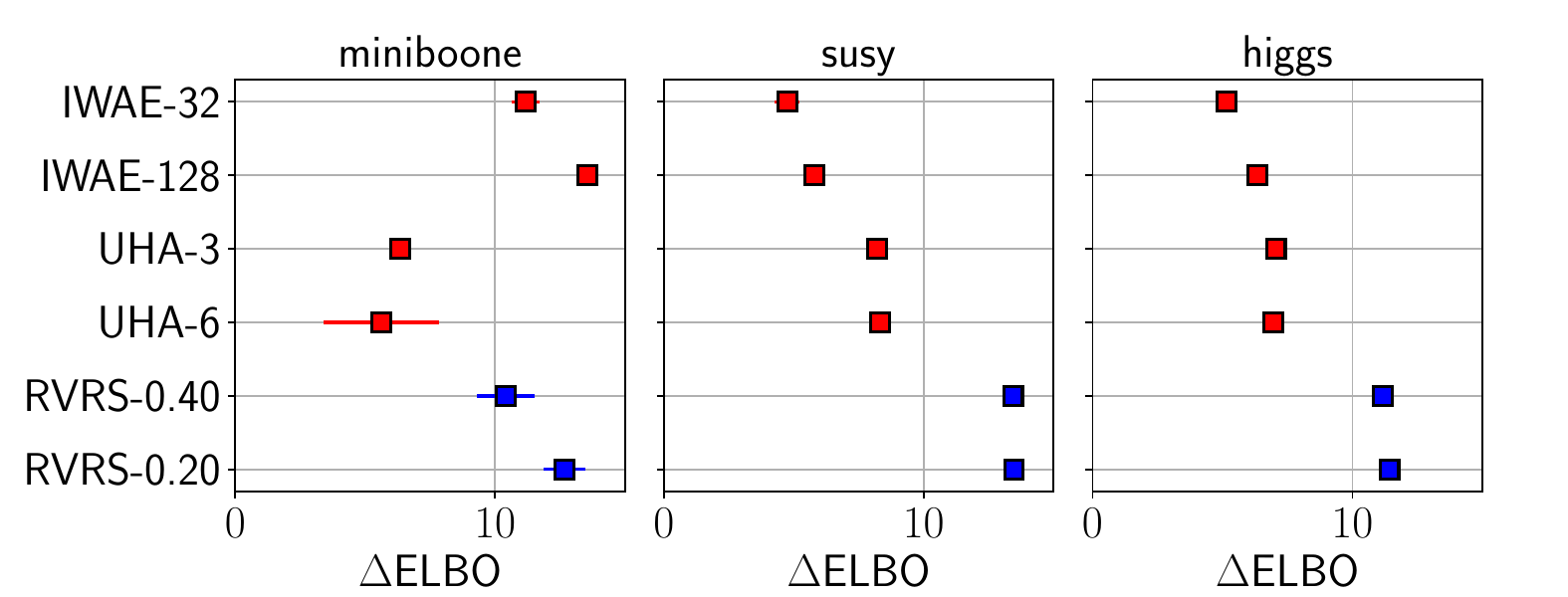}
    \caption{ 
    We depict ELBO improvements above the multivariate Normal baseline for $6$ variational methods on $3$
    Gaussian process classification tasks. 
    Results are averaged across five runs and error bars denote two standard deviations.
    }
    \label{fig:gp}
\end{center}
\end{figure}

To probe the ability of RVRS to handle both model learning and higher-dimensional latent spaces, we
consider Gaussian process models for binary classification. For each of $3$ datasets we consider $N=256$
data points and thus $D=256$ latent dimensions. Model parameter $\bth$ are the $D_\bx+1$ kernel hyperparameters,
where $D_\bx$ is the dimension of the inputs with $D_\bx \in \{18, 28, 51\}$. See \figref{fig:gp} for
the results. Perhaps surprisingly given the large dimension, we find that RVRS matches or exceeds the
performance of the other methods. This is even true for IWAE with $K=128$ particles.
Notably UHA-$6$ does about the same as UHA-$3$, emphasizing the difficulty of optimizing the UHA 
ELBO---which effectively differentiates through a short MCMC chain---with its potential for numerical instability w.r.t.~the
step size and mass matrix that define the Hamiltonian dynamics. 
Thus although (high-dimensional) gradients offer a lot of information about the posterior density, 
effectively utilizing that information can be challenging to the point where rejection sampling---which we would
generally expect to be less effective for large $D$---can be just as effective or even more so. 

%% SUSY (D=18)
%% Higgs (D=28)
%% MiniBooNE (D=51)

%%%%%%%%%%%%%%%%%%%%
%%%%%%%%%%%%%%%%%%%%

%%%%%%%%%%%%%%%%%%%%%%%%%%%%%%%%%%%%%%%%%%%%%%%%%%
\subsection{Variational autoencoders}
\label{sec:vae}
%%%%%%%%%%%%%%%%%%%%%%%%%%%%%%%%%%%%%%%%%%%%%%%%%%

We compare conventional ELBO training with RVRS, IWAE, and UHA/DAIS on a VAE \citep{kingma2013auto}
trained on statically binarized MNIST. We use the same encoder-decoder architecture as in \citep{burda2015importance}
and set the dimension of the latent variable to $D=50$. 
See Table~\ref{table:vae} for the results.
Notably RVRS is faster than IWAE because RVRS only requires computing gradients through a small number ($S=2$) of accepted samples. 
We find that RVRS consistently outperforms IWAE but is edged out by UHA with many gradient steps. The good performance
of UHA with $K=20$ gradient steps comes at significant computational cost, however, as training is $\sim5.6$x slower than RVRS-$0.025$.
Thus an attractive feature of RVRS trained with Algorithm \ref{alg:semibiased} is that it can make effective use of parallel hardware, 
while UHA is bottlenecked by the serial nature of MCMC chains.

\begin{table}[H]
	\centering
	\resizebox{1.02\textwidth}{!}{%
		\begin{tabular}{l||l|l|l|l|l|l|l|l|l}
			\hline
			\small{Method} & \small{Standard VAE} & \small{IWAE-10} & \small{IWAE-20} & \small{IWAE-40} & \small{UHA-$10$} & \small{UHA-$20$} &  \small{RVRS-$0.1$} & \small{RVRS-$0.05$} & \small{RVRS-$0.025$}  \\ \hline\hline 
			
			%\small{Train -ELBO}  & \small{$92.00 \pm 0.10$} & \small{$88.66 \pm 0.08$} & \small{$87.93 \pm 0.03$} & \small{$87.35 \pm 0.08$} & \small{$87.01 \pm 0.08$} & \small{$86.04 \pm 0.33$} & \small{$87.58 \pm 0.08$} & \small{$87.10 \pm 0.07$} & \small{$86.87 \pm 0.06$} \\ \hline
			
			%\small{Train ELBO}  & \small{$-92.00 \pm 0.10$} & \small{$-88.66 \pm 0.08$} & \small{$-87.93 \pm 0.03$} & \small{$-87.35 \pm 0.08$} & \small{$-87.01 \pm 0.08$} & \small{$-86.04 \pm 0.33$} & \small{$-87.58 \pm 0.08$} & \small{$-87.10 \pm 0.07$} & \small{$-86.87 \pm 0.06$} \\ \hline
			
			%\small{Test -ELBO}  & \small{$95.30 \pm 0.14$} & \small{$91.21 \pm 0.07$} & \small{$90.44 \pm 0.07$} & \small{$89.81 \pm 0.09$} & \small{$89.75 \pm 0.09$} & \small{$88.46 \pm 0.22$} & \small{$90.74 \pm 0.16$} & \small{$90.00 \pm 0.12$} & \small{$89.55 \pm 0.12$} \\ \hline
			
			\small{$-$ELBO} & \small{$95.30 \pm 0.14$} & \small{$91.21 \pm 0.07$} & \small{$90.44 \pm 0.07$} & \small{$89.81 \pm 0.09$} & \small{$89.75 \pm 0.09$} & \small{$88.46 \pm 0.22$} & \small{$90.74 \pm 0.16$} & \small{$90.00 \pm 0.12$} & \small{$89.55 \pm 0.12$} \\ \hline
			
			\small{ms / grad} & \small{$0.70$} & \small{$1.06$} & \small{$1.49$} & \small{$1.97$} & \small{$5.17$} & \small{$9.73$} &  \small{$1.19$}&  \small{$1.36$} & \small{$1.75$} \\ \hline
		\end{tabular}
	} % end resize
	\vspace{2mm}
	\caption{ 
		We report negative ELBO objectives (lower is better; mean $\pm$ standard deviation over $5$ replicates) 
        computed on held-out test data
        together with gradient step times for the VAE experiment in \secref{sec:vae}. 
        In all cases we report results using the same objective used during training.
        Results are obtained with a RTX 2070 GPU.}
	\label{table:vae}
\end{table}

%%%%%%%%%%%%%%%%%%%%%%%%%%%%%%%%%%%%%%%%%%%%%%%%%%
\subsection{Hierarchical modeling}
\label{sec:local}
%%%%%%%%%%%%%%%%%%%%%%%%%%%%%%%%%%%%%%%%%%%%%%%%%%

We evaluate RVRS on a hierarchical model with local latent variables that was also
considered by \citet{jankowiak2022surrogate}.
In detail we consider a Bayesian linear regressor that utilizes a Student’s t likelihood.
Since this likelihood can be represented as a continuous mixture of Normal distributions, this
choice corresponds to a hierarchical model with local Gamma latent variables that can be integrated out exactly.
We compare three variational approaches, all of which
use a mean-field Normal distribution for the global latent coefficient.
We consider two semi-parametric approaches---Semi-DAIS and Semi-RVRS---that 
only differ in how the approximate posterior over the local latent variables is contructed.
We also compare against an oracle baseline obtained by integrating out the
Gamma variates before performing variational inference. 
This oracle represents an upper performance bound
on the two semi-parametric approaches.
See Table~\ref{table:local} for results.\footnote{
Note that the Semi-RVRS results in Table~\ref{table:local} were obtained using the unbiased 
sampler Algorithm~\ref{alg:semiunbiased} during training.  In \figref{fig:biasedsemi} in \secref{sec:addexp} we 
provide a comparison to results obtained with the faster biased sampler defined in Algorithm~\ref{alg:semibiased}, 
which yields nearly identical performance.}

For both datasets we find that Semi-RVRS nearly matches the performance of the oracle, even with $\ZZ_\target = 0.5$,
implying that Semi-RVRS yields a conditional posterior over the local latent variables that is nearly exact.
Semi-DAIS also yields good performance but fails to approach the oracle upper bound even with $K=32$ gradient steps.
%\footnote{For these two datasets the time per gradient step of Semi-RVRS-$0.50$ roughly 
%equals that of Semi-DAIS-$16$.}
Thus this experiment highlights a particular strength of RVRS, namely dealing with low-dimensional latent variables, 
a regime in which methods based on expensive gradient-based MCMC can be overkill. Perhaps remarkably RVRS can still be competive
in higher dimensions, as demonstrated in the Gaussian process experiment in \secref{sec:gp}.

%%%%%%%%%%%%%%%%
\begin{table}[H]
    \centering
    \resizebox{.99\textwidth}{!}{%
\begin{tabular}{l||l|l|l|l|l|l}
\hline
\small{Dataset} & \small{Semi-DAIS-$8$} & \small{Semi-DAIS-$16$} & \small{Semi-DAIS-$32$}   & \small{Semi-RVRS-$0.50$} & \small{Semi-RVRS-$0.10$}  & \small{Oracle}   \\ \hline\hline 
\small{Pol}  & \small{$37.8 \pm 1.6$} & \small{$73.8 \pm 1.6$} & \small{$99.8 \pm 1.6$} & \small{$116.9 \pm 0.9$} & \small{$140.2 \pm 0.9$} & \small{$143.1 \pm 0.4$} \\ \hline
\small{Bike} & \small{$49.5 \pm 0.8$} & \small{$110.8 \pm 0.7$} & \small{$155.8 \pm 0.8$} & \small{$187.8 \pm 0.7$} & \small{$223.5 \pm 0.7$} &  \small{$228.6 \pm 0.5$} \\ \hline
\end{tabular}
     } % end resize
\vspace{2mm}
    \caption{ 
        We report ELBO improvement above a mean-field baseline for the hierarchical model in \secref{sec:local} (mean $\pm$
        standard deviation).
    Results are averaged across 5 replicates.}
\label{table:local}
\end{table}
%%%%%%%%%%%%%%

%*** bike ***
%[SemiRVRS-10]	  \small{$223.5 \pm 0.7$}
%[SemiRVRS-50]	  \small{$187.8 \pm 0.7$}
%[SemiDAIS-8]	  \small{$49.5 \pm 0.8$}
%[SemiDAIS-16]	  \small{$110.8 \pm 0.7$}
%[ORACLE]	  \small{$228.6 \pm 0.5$}
%[SemiDAIS-32]	  \small{$155.8 \pm 0.8$}
%*** pol ***
%[SemiRVRS-10]	  \small{$140.2 \pm 0.9$}
%[SemiRVRS-50]	  \small{$116.9 \pm 0.9$}
%[SemiDAIS-8]	  \small{$37.8 \pm 1.6$}
%[SemiDAIS-16]	  \small{$73.8 \pm 1.6$}
%[ORACLE]	  \small{$143.1 \pm 0.4$}
%[SemiDAIS-32]	  \small{$99.8 \pm 1.6$}

%%%%%%%%%%%%%%%%%%%%%%%%%%%%%%%%%%%%%
\section{Discussion}
\label{sec:disc}
%%%%%%%%%%%%%%%%%%%%%%%%%%%%%%%%%%%%%

Given its relative simplicity, it is remarkable that RVRS can match---and in some cases exceed---the performance
of more complex gradient-based hybrid variational inference schemes like UHA/DAIS \citep{geffner2021mcmc,zhang2021differentiable}.
For this reason we believe it could be especially valuable to combine RVRS with other methods,
since RVRS with moderate $\ZZ_\target$ provides a relatively cheap and simple way to achieve non-trivial refinement of the
proposal distribution $\qphi(\bz)$. For example it would be natural to use a normalizing flow or 
MixFlow \citep{xu2023mixflows} in place of a simple parametric proposal.
This could be especially attractive in cases where there are diminishing returns to e.g.~using more layers (in the case of normalizing 
flows) or more pushforwards (in the case of MixFlows).
Importantly in RVRS we only need to differentiate through \emph{accepted} samples $\bz \sim \rr$, which limits
the computational cost of leveraging RVRS. More broadly the design space of hybrid variational algorithms remains
only partially explored and involves various algorithmic and computational trade-offs. 
As such we expect that RVRS could be a useful
component in the design of future hybrid variational inference methods.

\begin{ack}
    We warmly thank Matthew D.~Hoffman for feedback on a draft manuscript.
    MJ's contributions to the work reported here are independent of his role at Generate Biomedicines.
\end{ack}

%\section*{References}

%%%%%%%%%%%%%%%%%%%%%%%%%%%%%%%%%%%%%%%%%%%%%%%%%%%%%%%%%%%%
\bibliographystyle{plainnat}
\bibliography{ref}
%%%%%%%%%%%%%%%%%%%%%%%%%%%%%%%%%%%%%%%%%%%%%%%%%%%%%%%%%%%%

%%%%%%%%%%%%%%
\newpage
\appendix
%%%%%%%%%%%%%%

%%%%%%%%%%%%%%%%%%%%%%%%%%%%%%%%%%%%%
\section{Gradient estimator for the parameters of the proposal distribution}
\label{sec:phiest}
%%%%%%%%%%%%%%%%%%%%%%%%%%%%%%%%%%%%%

%%%%%%%%%%%%%%%%%%%%%%%%%%%%%%%%%%%%
\subsection{Reparameterized $\phi$ gradient estimator}
\label{sec:phigrad}
%%%%%%%%%%%%%%%%%%%%%%%%%%%%%%%%%%%%

The covariance in \eqnref{eqn:phigrad} can be converted into a pathwise gradient estimator.
To see this consider the ``fundamental pathwise gradient identity'' (see e.g.~\cite{jankowiak2018pathwise,mohamed2020monte})\footnote{
Note that \eqnref{eqn:pathidentity} is equal to $\nabla_\bphi \EE_{q_\bphi(\bz)} \left[ f(\bz) \right] - \EE_{q_\bphi(\bz)} \left[ \nabla_\bphi  f(\bz) \right]$ but this fact is not needed for our derivation.} 
%%%
\begin{align}
\label{eqn:pathidentity}
 \EE_{q_\bphi(\bz)} \left[ f(\bz) \nabla_{\bphi} \log q_\bphi(\bz)  \right] = 
 \EE_{q_\bphi(\bz)} \left[ \frac{\partial f(\bz)}{\partial \bz} \cdot \nabla_{\bphi} \bz  \right] 
\end{align}
%%%
where $f(\bz)$ can depend on $\bphi$ and where $\nabla_{\bphi} \bz$ is a velocity field for the parameter $\bphi$
that can be derived via e.g.~the reparameterization trick if $q_\bphi(\bz)$ is reparameterizable.
Then use \eqnref{eqn:pathidentity} to derive the identity
%%%
\begin{align}
\label{eqn:repidentity}
 \EE_{\rr(\bz)} \left[ f(\bz) \nabla_{\bphi} \log q_\bphi(\bz)  \right] &= 
  \EE_{\qphi(\bz)} \left[  \frac{a_{\bth, \bphi}(\bz)}{\ZZr} f(\bz) \nabla_{\bphi} \log q_\bphi(\bz)  \right] \\
&=  \EE_{\qphi(\bz)} \left[  \frac{\partial}{\partial \bz} \left( \frac{a_{\bth, \bphi}(\bz)}{\ZZr} f(\bz) \right) \cdot  \nabla_{\bphi} \bz \right] \\
&=  \EE_{\qphi(\bz)} \left[  \frac{1}{\ZZr}   \frac{\partial}{\partial \bz} \left( a_{\bth, \bphi}(\bz) f(\bz)  \right) \cdot  \nabla_{\bphi} \bz \right] \\
&=  \EE_{\qphi(\bz)} \left[  \frac{a_{\bth, \bphi}(\bz)}{\ZZr} \left( 
f(\bz) \frac{\partial \log a_{\bth, \bphi}(\bz) }{\partial \bz} + \frac{\partial f(\bz) }{\partial \bz} \right)   \cdot  \nabla_{\bphi} \bz \right] \\
&=  \EE_{\rr(\bz)} \left[  \left( 
\label{eqn:finalrepidentity}
f(\bz) \frac{\partial \log a_{\bth, \bphi}(\bz) }{\partial \bz} + \frac{\partial f(\bz) }{\partial \bz} \right)   \cdot  \nabla_{\bphi} \bz \right] %\\ 
%&=  \EE_{\rr(\bz)} \left[  \frac{\partial}{\partial \bz}  \left( 
%\overline{f_\bphi(\bz)}  \log a_{\bth, \bphi}(\bz) + f_\bphi(\bz) \right)   \cdot  \nabla_{\bphi} \bz \right]
\end{align}
%%%
%where the bar in $\overline{f_\bphi(\bz)}$ denotes the $\texttt{stop\_gradient}$ op (in the language of JAX) a.k.a.~the $\texttt{detach}$ op (in the language of PyTorch).
If we make the substitution $f(\bz) \rightarrow g(\bz) a_{\bth, \bphi}(\bz)$ in \eqnref{eqn:finalrepidentity} this identity can be re-expressed as
%%%
\begin{align}
\label{eqn:repidentity2}
 \EE_{\rr(\bz)} \left[g(\bz) a_{\bth, \bphi}(\bz) \nabla_{\bphi} \log q_\bphi(\bz)  \right] &=  \EE_{\rr(\bz)} \left[  \left( 
2 g(\bz) \frac{\partial a_{\bth, \bphi}(\bz) }{\partial \bz} + a_{\bth, \bphi}(\bz)  \frac{\partial g(\bz) }{\partial \bz} \right)   \!\cdot\!  \nabla_{\bphi} \bz \right] %\\ \nonumber
%&= \EE_{\rr(\bz)} \left[  \left( 
%2 g(\bz) a_{\bth, \bphi}(\bz) (1-a_{\bth, \bphi}(\bz) ) + a_{\bth, \bphi}(\bz)  \frac{\partial g(\bz) }{\partial \bz} \right)   \cdot  \nabla_{\bphi} \bz \right] 
\end{align}
%%%
Using the final form of the identity \eqnref{eqn:repidentity2} we can rewrite \eqnref{eqn:phigrad} as follows:
%%%
\begin{align}
\label{eqn:supp:repphigrad}
\nabla_\bphi \elbo &= {\rm COV}_{\rr(\bz)} \left[ \AAA(\bz), a_{\bth, \bphi}(\bz) \nabla_\bphi \log q_\bphi(\bz)  \right]  \\ \nonumber
&= \EE_{\rr(\bz)} \left[ \; \AAAbar(\bz) 
a_{\bth, \bphi}(\bz) \nabla_\bphi \log q_\bphi(\bz)  \right]  \\ \nonumber
&= \EE_{\rr(\bz)} \left[  \left( 2 \AAAbar(\bz)   \frac{\partial a_{\bth, \bphi}(\bz) }{\partial \bz}  +  a_{\bth, \bphi}(\bz) \frac{\partial \AAA(\bz)  }{\partial \bz}
 \right) \cdot \nabla_{\bphi} \bz \right] 
\end{align}
%%%
where we have defined
%%%
\begin{align}
    \label{eqn:aaabar}
\AAAbar(\bz) \equiv  \AAA(\bz) - \EE_{\rr(\bz^\prime)}\left[ \AAA(\bz^\prime)\right]
\end{align}
%%%
and used that 
%%%
\begin{align}
\frac{\partial }{\partial \bz} \AAAbar(\bz) = \frac{\partial }{\partial \bz} \AAA(\bz) 
\end{align}
%%%
We also note that \eqnref{eqn:supp:repphigrad} can be expressed in covariance form as follows (although
we prefer the more compact form utilizing $\AAAbar(\bz)$):
%%%
\begin{align}
    \nabla_\bphi \elbo = {\rm COV}_{\rr(\bz)} \left[  2 \AAA(\bz), \frac{\partial a_{\bth, \bphi}(\bz) }{\partial \bz} \cdot \nabla_{\bphi} \bz \right] +
    \EE_{\rr(\bz)} \left[ a_{\bth, \bphi}(\bz) \frac{\partial \AAA(\bz)  }{\partial \bz}
 \cdot \nabla_{\bphi} \bz \right]
\end{align}
%%%
Finally we note that, as we would expect, \eqnref{eqn:supp:repphigrad} reduces to the standard reparameterized gradient
in the limit that $a_{\bth, \bphi}(\bz)  \rightarrow 1$ and $\rr(\bz) \rightarrow \qphi(\bz)$:
%%%
\begin{align}
\nabla_\bphi \elbo \rightarrow \EE_{\qphi(\bz)} \left[  \frac{\partial \AAA(\bz)  }{\partial \bz} \cdot \nabla_{\bphi} \bz \right]  = \EE_{\qphi(\bz)} \left[  \frac{\partial }{\partial \bz} \left(  \log \pth(\bx, \bz) -  \log q_\bphi(\bz) \right) \cdot \nabla_{\bphi} \bz \right] 
\end{align}
%%%

%%%%%%%%%%%%%%%%%%%%%%%%%%%%%%%%%%%%
\subsection{Automatic differentation and Monte Carlo details for ELBO and gradient estimation}
\label{sec:rvrsautogradsupp}
%%%%%%%%%%%%%%%%%%%%%%%%%%%%%%%%%%%%

To get unbiased estimates of \eqnref{eqn:repphigrad} we need\footnote{Note that another option would be to keep
a running estimate of $\EE_{\rr(\bz)}\left[ \AAA(\bz)\right]$ and use this in \eqnref{eqn:supp:repphigrad} and 
\eqnref{eqn:aaabar}. This would result in a biased estimator, but the bias should be minimal given that $\bphi$ and $\bth$ change slowly over the course of optimization. This is an interesting option that can reduce computational cost by opening the door to single-sample (i.e.~$S=1$) gradient estimation. While we do not explore this option empirically, we have every reason to expect that it would work well.} to draw $S>1$ samples simultaneously, i.e.~just like VRS RVRS utilizes a multi-sample 
objective.  In particular if $\bz_s \sim \rr$ for $s=1,...,S$ and we use a reparameterized
sampler for $q_\bphi(\bz)$ so that $\bz_s$ depends explicitly on $\bphi$ according to the automatic
differentiation system (e.g.~\texttt{torch.autograd}), we can define the following surrogate ELBO:
%%%
\begin{align}
\label{eqn:mcphigrad}
\LL_{\rm surr} = \frac{2}{S - 1} \sum_{s=1}^S  \reallywidetilde{ \Big  \{ \AAA(\bz_s) - \mu_\AAA(\bz_{1:S}) \Big \}} \reallywidetilde{a_{\bth, \bphi}}(\bz_s) +
 \frac{1}{S} \sum_{s=1}^S  \reallywidetilde{a_{\bth, \bphi}(\bz_s)} \reallywidetilde{\AAA}(\bz_s)
 \end{align}
%%%
where $\reallywidetilde{{\rm f(\bz)}}$ denotes $\texttt{stop\_gradient}(f(\bz))$ and
$\reallywidetilde{{\rm f}}(\bz)$ denotes $\texttt{stop\_gradient}(f)(\bz)$ and
%%%
\begin{align}
\mu_\AAA(\bz_{1:S}) \equiv \frac{1}{S} \sum_{s=1}^S \AAA(\bz_s)  
\end{align} 
%%%
To derive \eqnref{eqn:mcphigrad} we used the identity
%%%
\begin{align}
    \label{eqn:covidentity}
{\rm COV}_{r(\bz)}[A(\bz), B(\bz)] &\approx \tfrac{1}{S } \sum_{s=1}^S \left(A(\bz_s) - \tfrac{1}{S - 1}\sum_{s^\prime \ne s} A(\bz_{s^\prime}) \right) B(\bz_s) \\
&=  \tfrac{1}{S } \sum_{s=1}^S \left(A(\bz_s) - \tfrac{1}{S - 1} \left(-A(\bz_s) +  \sum_{s^\prime=1}^S A(\bz_{s^\prime}) \right) \right) B(\bz_s) \\
&=  \tfrac{1}{S } \sum_{s=1}^S \left((1 + \tfrac{1}{S-1}) A(\bz_s) - \tfrac{1}{S - 1} \sum_{s^\prime=1}^S A(\bz_{s^\prime})  \right) B(\bz_s) \\
&=  \tfrac{1}{S } \sum_{s=1}^S \left(\tfrac{S}{S-1} A(\bz_s) - \tfrac{S}{S - 1} \tfrac{1}{S} \sum_{s^\prime=1}^S A(\bz_{s^\prime}) \right) B(\bz_s) \\
&=  \tfrac{1}{S-1 } \sum_{s=1}^S \left( A(\bz_s) - \tfrac{1}{S} \sum_{s^\prime=1}^S A(\bz_{s^\prime}) \right) B(\bz_s) \\
\end{align} 
%%%
 By construction when $\LL_{\rm surr}$ in \eqnref{eqn:mcphigrad} is run through autograd we get an unbiased
 estimate of \eqnref{eqn:repphigrad}. For the purposes of tracking the ELBO for evaluation we get a (biased) MC estimator as follows:
%%%
\begin{align}
\LL = \frac{1}{S} \sum_{s=1}^S \AAA(\bz_s) + \log \ZZr 
 \end{align}
%%%
where
%%%
\begin{align}
 \log \ZZr = \log \EE_{q_\bphi(\bz)} \left[a_{\bth, \bphi}(\bz) \right] \approx \log \frac{1}{S} \sum_{s=1}^Sa_{\bth, \bphi}(\bz_s^\prime)
 \end{align}
%%%
where $\bz_s^\prime \sim \qphi$ for $s=1,...,S$. In practice we use a large number of samples (e.g.~$S\sim10^4-10^5$)
to evaluate $\log \ZZr$.

%%%%%%%%%%%%%%%%%%%%%%%%%%%%%%%%%%%%
\subsection{Runtime considerations}
%%%%%%%%%%%%%%%%%%%%%%%%%%%%%%%%%%%%

Nothing about VRS or RVRS depends on the specific ansatz for $\aaa(\bz)$ in \eqnref{eqn:rdef}, apart
from the generic requirement (for RVRS) that $\aaa(\bz)$ depend on $\bphi$ through $\qphi(\bz)$
and that $\aaa(\bz) \in [0, 1]$.
We can thus consider other forms of $\aaa(\bz)$. One potential problem with
$\aaa(\bz) =\sigma(\log \pth(\bx, \bz) - \log \qphi(\bz) + T)$ is that it can lead to very small
acceptance probabilities if $T$ is poorly adapted. Consequently it can be useful to place guardrails that
mitigate against this possibility. In the following we consider the simple ansatz

%%%
\begin{align}
    \label{eqn:aaaeps}
    \aaaeps(\bz) =\epsilon + (1-\epsilon)\aaa(\bz)=\epsilon + (1-\epsilon)\sigma(\log \pth(\bx, \bz) - \log \qphi(\bz) + T)
 \end{align}
%%%
 where $\epsilon>0$ is some small fixed constant like $\epsilon =10 ^{-3}$ or $\epsilon =10 ^{-2}$.
With this choice $\aaaeps(\bz) \in (\epsilon, 1)$ which guarantees that
$\ZZ_r \ge \epsilon$. Although this shouldn't be necessary if sufficient care is taken with $T$ adaptation,
we use the ansatz in \eqnref{eqn:aaaeps} in all our experiments to guard against the possibility of excessive runtimes. 
Here we describe how this choice modifies Prop.~\ref{prop:gradest}.

We begin with the VRS formula
%%%
\begin{align}
    \nabla_\bphi \elbo = {\rm COV}_{\rreps(\bz)} \left[ \AAA(\bz), \nabla_\bphi \log \{ \qphi(\bz) \aaaeps(\bz) \}\right]
\end{align}
%%%
In the limit that $\epsilon=0$ this simplifies to 
${\rm COV}_{\rr(\bz)} \left[ \AAA(\bz), \aaa(\bz) \nabla_\bphi \log \qphi(\bz)\right]$,
see \eqnref{eqn:phigradsupp}. A bit more algebra is involved if $\epsilon>0$.
Indeed we have
%%%
\begin{align}
    \nabla_\bphi \log \aaaeps(\bz) = \frac{(1-\epsilon)\nabla_\bphi \aaa(\bz)}{\epsilon + (1-\epsilon)\aaa(\bz)} 
    = \frac{\nabla_\bphi \log \aaa(\bz)}{\frac{\epsilon}{\aaa(\bz)(1 - \epsilon)} + 1} 
\end{align}
%%%
Since $\nabla_\bphi \log \aaa(\bz) = (\aaa(\bz) -1) \nabla_\bphi \log \qphi(\bz)$ we can write
%%%
\begin{align}
    \nabla_\bphi \log \{\qphi(\bz) \aaaeps(\bz)\} &=\left(1 + \frac{\aaa(\bz)-1}{\frac{\epsilon}{\aaa(\bz)(1 - \epsilon)} + 1}\right)\nabla_\bphi \log \qphi(\bz)  \\
    &= \frac{\zeta + \aaa(\bz)^2}{\zeta + \aaa(\bz)}\nabla_\bphi \log \qphi(\bz)
\end{align}
%%%
where we have defined $\zeta \equiv \epsilon / (1 - \epsilon)$. Thus we have
%%%
\begin{align}
    \nabla_\bphi \elbo &= \EE_{\rreps(\bz)} \left[ \AAAbar(\bz) \nabla_\bphi \log \{ \qphi(\bz) \aaaeps(\bz) \}\right] \\
    &= \EE_{\rreps(\bz)} \left[ \AAAbar(\bz) \frac{\zeta + \aaa(\bz)^2}{\zeta + \aaa(\bz)}\nabla_\bphi \log \qphi(\bz) \right]
\end{align}
%%%
We can now appeal to the same logic in \eqnref{eqn:repidentity} with
%%%
\begin{align}
f(\bz) \rightarrow \AAAbar(\bz) \frac{\zeta + \aaa(\bz)^2}{\zeta + \aaa(\bz)} 
\end{align}
%%%
to write
%%%
\begin{align}
\nabla_\bphi \elbo = \EE_{\rreps(\bz)} \left[  \left(
f(\bz) \frac{\partial \log \aaaeps(\bz) }{\partial \bz} + \frac{\partial f(\bz) }{\partial \bz} \right)   \cdot  \nabla_{\bphi} \bz \right]
    \label{eqn:epsphigrad}
\end{align}
%%%
We can then use \eqnref{eqn:epsphigrad} to construct a Monte Carlo surrogate ELBO estimator like in
\secref{sec:rvrsautogradsupp}, though we spare the reader the tedious derivation.
The upshot is the following estimator:
%%%
\begin{align}
\label{eqn:mcphigradeps}
    \LL_{\rm surr, \epsilon} = &\frac{1}{S - 1} \sum_{s=1}^S  \reallywidetilde{ \Big  \{ \AAA(\bz_s) - \mu_\AAA(\bz_{1:S}) \Big \}} 
    \left(\reallywidetilde{\frac{\zeta + \aaa(\bz)^2}{\zeta + \aaa(\bz)}} \reallywidetilde{\log \aaaeps}(\bz_s) + 
    \frac{\zeta + \reallywidetilde{\aaa}(\bz)^2}{\zeta + \reallywidetilde{\aaa}(\bz)}
    \right) + \nonumber\\
    &\frac{1}{S} \sum_{s=1}^S  \reallywidetilde{\frac{\zeta + \aaa(\bz)^2}{\zeta + \aaa(\bz)}} \reallywidetilde{\AAA}(\bz_s)
 \end{align}
%%%
It is straightforward to check that this reduces to \eqnref{eqn:mcphigrad} when $\epsilon=\zeta=0$.

%%%%%%%%%%%%%%%%%%%%%%%%%%%%%%%%%%%%%
\section{Proof of proposition 2}
\label{sec:globalproof}
%%%%%%%%%%%%%%%%%%%%%%%%%%%%%%%%%%%%%

We want to bound the variational gap $\Delta$ between $ \log \pth(\bx) \equiv \log \EE_{\pth(\bz)} \left [ \pth(\bx|\bz) \right ]$
and the ELBO
%%%
\begin{align}
\Delta = \log \pth(\bx) - {\rm ELBO} 
\end{align}
%%%
as a function of $T$. We work under the assumption that $\qphi(\bz)$ is sufficiently heavy-tailed so that 
the ratio $\frac{\pth(\bx,\bz) }{\qphi(\bz)}$ is well-behaved (see \eqnref{eqn:momentcond} below for the precision condition).

We have
%%%
\begin{align}
\label{eqn:deltadefn}
\Delta = {\rm KL}(\rr(\bz) || \pth(\bz|\bx) ) 
= \EE_{\rr(\bz)} \left[ \log \frac{\rr(\bz)}{\pth(\bz|\bx) } \right]\ge 0
\end{align}
%%%
The KL divergence in \eqnref{eqn:deltadefn} can be decomposed into a positive contribution from where the logarithm is positive
and a negative contribution from where the logarithm is negative. 
Since the KL divergence is non-negative the magnitude of the positive contribution is larger than or equal
to the magnitude of the negative contribution. Consequently to bound $\Delta$ it suffices to bound
the positive contribution. 

The ratio in the log in \eqnref{eqn:deltadefn} is given by
%%%
\begin{align}
\frac{\rr(\bz)}{\pth(\bz|\bx) } &= \frac{\qphi(\bz)}{\ZZr \pth(\bz|\bx) } a_{\bth, \bphi}(\bz)
= \frac{\qphi(\bz)}{\ZZr \pth(\bz|\bx) }  \frac{1}{1 + e^{-T} \frac{\qphi(\bz)}{\ZZp \pth(\bz|\bx)}} \\ 
&=  \frac{1}{\frac{\ZZr \pth(\bz|\bx) }{\qphi(\bz)}  + e^{-T} \frac{\ZZr}{\ZZp}} = \frac{1}{1 + f(\bz|T)}
\end{align}
%%%
where $\ZZp \pth(\bz|\bx) = \pth(\bx,\bz)$ with $\ZZp \equiv \pth(\bx)$ and
%%%
\begin{align}
f(\bz|T) \equiv e^{-T} \frac{\ZZr}{\ZZp} - 1 + \ZZr \frac{\pth(\bz|\bx)}{q_\bphi(\bz)} 
\end{align}
%%%
Thus $\log \frac{\rr(\bz)}{\pth(\bz|\bx) } > 0$ implies that $f(\bz|T) < 0$ so that our task is
to bound $f(\bz|T)$ from below.
Since $ \ZZr \frac{\pth(\bz|\bx)}{q_\bphi(\bz)}  > 0$ we have that
%%%
\begin{align}
f(\bz|T) > e^{-T} \frac{\ZZr}{\ZZp} - 1
\end{align}
%%%
We compute
%%%
\begin{align}
\ZZr = \EE_{q_\bphi(\bz)} \left[a_{\bth, \bphi}(\bz) \right] =
 \EE_{q_\bphi(\bz)} \left[ \frac{1}{1 + e^{-T} \frac{\qphi(\bz)}{\ZZp \pth(\bz|\bx)}}\right]
 \end{align}
%%%
so that
%%%
\begin{align}
\frac{e^{-T}}{\ZZp} \ZZr &=  \EE_{q_\bphi(\bz)} \left[ \frac{\frac{e^{-T}}{\ZZp} }{1 + e^{-T} \frac{\qphi(\bz)}{\ZZp \pth(\bz|\bx)}}\right] =
\EE_{q_\bphi(\bz)} \left[ \frac{1 }{e^T \ZZp+ \frac{\qphi(\bz)}{ \pth(\bz|\bx)}}\right]  \\
&=\EE_{q_\bphi(\bz)} \left[ \frac{1 }{\frac{\qphi(\bz)}{ \pth(\bz|\bx)} \left( 1 + e^T \ZZp \frac{ \pth(\bz|\bx)}{\qphi(\bz)} \right)}\right] 
=\EE_{\pth(\bz|\bx)} \left[ \frac{1 }{ 1 + e^T \ZZp \frac{ \pth(\bz|\bx)}{\qphi(\bz)} }\right]  
 \end{align}
%%%
and therefore
%%%
\begin{align}
\frac{e^{-T}}{\ZZp} \ZZr -1 =  
\EE_{\pth(\bz|\bx)} \left[ \frac{-e^T \ZZp \frac{ \pth(\bz|\bx)}{\qphi(\bz)}  }{ 1 + e^T \ZZp \frac{ \pth(\bz|\bx)}{\qphi(\bz)} }\right] 
> \EE_{\pth(\bz|\bx)} \left[ -e^T \ZZp \frac{ \pth(\bz|\bx)}{\qphi(\bz)}  \right]    = -e^T \xi
 \end{align}
%%%
where we have defined
%%%
\begin{align}
\label{eqn:momentcond}
\xi \equiv \EE_{\pth(\bz|\bx)} \left[  \frac{ \pth(\bx,\bz)}{\qphi(\bz)}  \right] =
    \EE_{\pth(\bz|\bx)} \left[  \frac{ \ZZp \pth(\bz|\bx)}{\qphi(\bz)}  \right] > 0 
 \end{align}
%%%
which is finite by assumption so that we can conclude
%%%
\begin{align}
f(\bz|T) > -e^T \xi
\end{align}
%%%
Since
%%%
\begin{align}
\log \frac{\rr(\bz)}{\pth(\bz|\bx) } = - \log (1 + f(\bz|T))
\end{align}
%%%
and
%%%
\begin{align}
    \label{eqn:logidentity}
- \log (1 - x) \le \frac{3}{2} x \qquad {\rm for} \qquad 0 \le x \le \frac{1}{2}
\end{align}
%%%
we conclude that
%%%
\begin{align}
\log \frac{\rr(\bz)}{\pth(\bz|\bx) } < \frac{3}{2} e^T \xi  
\qquad {\rm for\; } \bz {\rm \;such\; that} \qquad\log \frac{\rr(\bz)}{\pth(\bz|\bx) }  \ge 0 \qquad {\rm and} \qquad  T < -\log 2 \xi
\end{align}
%%%
and consequently
%%%
\begin{align}
\Delta < \frac{3}{2} e^T \xi  \qquad {\rm for} \qquad T < -\log 2 \xi
\end{align} 
%%%
Since $e^T \rightarrow 0$ as $T \rightarrow -\infty$ we conclude that the variational
gap can be made arbitrarily tight. Of course the acceptance probability also goes to zero as $\sim e^T$ in this limit
so it becomes increasingly expensive to tighten the gap.

%%%%%%%%%%%%%%%%%%%%%%%%%%%%%%%%%%%%%%%
\subsection{Semi-RVRS: models with global and local latent variables}
%%%%%%%%%%%%%%%%%%%%%%%%%%%%%%%%%%%%%%%

Instead of considering generic unstructured models as above, we now consider the scenario introduced
in \secref{sec:hier}, i.e.~we consider models with both
a global latent variable $\bz_G$ and local latent variables $\{ \bz_n \}$, where $n=1,...,N$ indexes
the $N$ observed data points $\{ \bx_n \}$. (See \secref{sec:semisupp} for additional algorithmic details
on Semi-RVRS).
We assume the following conditional independence structure:
%%%
\begin{align}
    \pth(\bx_{1:N}, \bz_G, \bz_{1:N}) = \pth(\bz_G) \prod_{n=1}^N \pth(\bx_n | \bz_n, \bz_G) \pth(\bz_n | \bz_G)
\end{align}
%%%
We want to upper bound the variational gap, which is given by
%%%
\begin{align}
    \Delta = \KL \left( \qphi(\bz_G) \prod_n \rrn(\bz_n | \bz_G) \; \Big | \Big | \; \pth(\bz_G | \bx_{1:N}) \prod_n \pth(\bz_n | \bz_G, \bx_n) \right)
\end{align}
%%%
where we have exploited the assumed conditional independence structure to factorize the posterior.
We now appeal to the chain rule of KL divergences which reads
%%%
\begin{equation}
\KL( q(a, b) || p(a, b) ) = \KL( q(a) || p(a) ) + \EE_{q(a)} \left[ \KL(q(b | a) || p(b | a) ) \right]
\end{equation}
%%%
to obtain
%%%
\begin{align}
    \label{eqn:localklfact}
    \Delta &= \KL \left( \qphi(\bz_G) \Big | \Big | \; \pth(\bz_G | \bx_{1:N}) \right) +
    \EE_{\qphi(\bz_G)} \left[ \KL \left( \prod_n \rrn(\bz_n | \bz_G) \; \Big | \Big | \; \prod_n \pth(\bz_n | \bz_G, \bx_n) \right) \right] \nonumber \\
    &= \KL \left( \qphi(\bz_G) \Big | \Big | \; \pth(\bz_G | \bx_{1:N}) \right) +
    \sum_n \EE_{\qphi(\bz_G)} \left[ \KL \left( \rrn(\bz_n | \bz_G) \; \Big | \Big | \; \pth(\bz_n | \bz_G, \bx_n) \right) \right]
\end{align}
%%%
Note that each $\bz_n$ KL divergence in \eqnref{eqn:localklfact} is precisely equal to the variational gap of
a RVRS variational distribution targeting the distribution $\pth(\bz_n | \bz_G, \bx_n)$ so we can apply
the same bounding logic as above (in particular exploiting the linearity in $x$ of the inequality in \eqnref{eqn:logidentity}) to each latent variable $\bz_n$ and obtain the following bound on the variational gap
%%%
\begin{equation}
    \Delta < \frac{3}{2} e^T \sum_{n=1}^N \xi_n + \KL \left( \qphi(\bz_G) \Big | \Big | \; \pth(\bz_G | \bx_{1:N}) \right)
\end{equation}
%%%
which is valid for $T < -\log 2 \max_n \xi_n$ where we assume that $T_n = T$ $\forall n$ and we define
%%%
\begin{equation}
    \xi_n \equiv \EE_{\qphi(\bz_G)} \EE_{\pth(\bz_n|\bz_G, \bx_n)} \left[  \frac{ \pth(\bx_n | \bz_n, \bz_G) \pth(\bz_n | \bz_G)}{\qphin(\bz_n)}  \right]
\end{equation}
%%%
Evidently this bound is only meaningful if all $\xi_n$ are finite, which will
be true if each proposal distribution $\qphin(\bz_n)$ is sufficiently heavy-tailed.

%%%%%%%%%%%%%%%%%%%%%%%%%%%%%%%%%%%%%
\section{Additional discussion of VRS}
\label{sec:vrssupp}
%%%%%%%%%%%%%%%%%%%%%%%%%%%%%%%%%%%%%

%%%%%%%%%%%%%%%%%%%%%%%%%%%%%%%%%
\subsection{Sampling cost}
\label{sec:samplingcost}
%%%%%%%%%%%%%%%%%%%%%%%%%%%%%%%%%

The number of proposal draws $\bz \sim \qphi(\cdot)$ generated before a sample is accepted
is governed by a geometric distribution with success probability $\ZZr \equiv \int \! d \bz \; \qphi(\bz) \aaa(\bz)$:
%%%%%%%%%%
\begin{equation}
\label{eqn:geomdist}
    {\rm Prob}(t^{\rm th}\; \text{sample accepted}) = \ZZr(1-\ZZr)^{t-1} \qquad {\rm with} \qquad t=1,2,...
\end{equation}
%%%%%%%%%%
Since the expected value of a geometric random variable is given by the reciprocal of the success probability, 
the expected number of draws from the proposal distribution is given by $\ZZr^{-1}$. Evidently, rejection
sampling becomes expensive for small $\ZZr$.

That the logic behind \eqref{eqn:geomdist} is correct can be corroborated by using the same
logic to compute the variational density $\rr(\bz)$ in terms of a geometric series:
%%%%%%%%%%
\begin{align}
    \rr(\bz) & = \sum_{t=1}^\infty {\rm Prob}\left(\text{accept} \; {\bm z} \; \text{at sampling step} \; t \Big \vert \text{rejected previous} \; t-1 \; \text{samples}\right) 
    {\rm Prob} \left(\text{reject} \; t-1 \; \text{samples}\right) \nonumber \\
    & = \sum_{t=1}^\infty \qphi(\bz) \aaa(\bz) \left(1-\int \qphi(\bz^\prime)\aaa(\bz^\prime) d\bz^\prime\right)^{t-1} \nonumber \\
    & = \qphi(\bz) \aaa(\bz) \sum_{t=0}^\infty (1-\ZZr)^t
     = \qphi(\bz) \aaa(\bz) \frac{1}{1-(1-\ZZr)} 
     = \frac{\qphi(\bz)\aaa(\bz)}{\ZZr}
\end{align}
%%%%%%%%%%
See \citet{bauer2019resampled} for an analogous derivation.

%%%%%%%%%%%%%%%%%%%%%%%%%%%%%%%%%
\subsection{Gradient estimators}
\label{sec:vrsgradestsupp}
%%%%%%%%%%%%%%%%%%%%%%%%%%%%%%%%%

The gradient estimator for proposal parameters $\bphi$ for the VRS ELBO can be expressed in a number
of equivalent ways
%%%
\begin{align}
\label{eqn:phigradsupp}
\nabla_\bphi \elbo 
    &= {\rm COV}_{\rr(\bz)} \left[ \AAA(\bz), \nabla_\bphi \log \{ \qphi(\bz) \aaa(\bz) \}\right] \\ \nonumber
 &= {\rm COV}_{\rr(\bz)} \left[ \AAA(\bz), (1 - \sigma(\ellT(\bz))) \nabla_\bphi \log \qphi(\bz)\right] \\ \nonumber
 &= {\rm COV}_{\rr(\bz)} \left[ \AAA(\bz), \sigma(-\ellT(\bz)) \nabla_\bphi \log \qphi(\bz) \right] \\ \nonumber
 &= {\rm COV}_{\rr(\bz)} \left[ \AAA(\bz), \aaa(\bz) \nabla_\bphi \log \qphi(\bz) \right]
\end{align}
%%%
where $\AAA(\bz) \equiv \log \pth(\bx, \bz) - \log \qphi(\bz) - \log \aaa(\bz)$.
In the limit that $T\rightarrow\infty$ we have $\aaa(\bz) \rightarrow 1$ and $\rr(\bz) \rightarrow \qphi(\bz)$.
Thus in this limit \eqnref{eqn:phigradsupp} becomes
%%%
\begin{align}
\label{eqn:phigradsupp2}
    \nabla_\bphi \elbo \rightarrow {\rm COV}_{\qphi(\bz)} \left[\log \pth(\bx, \bz) - \log \qphi(\bz) , \nabla_\bphi \log \qphi(\bz)\right] 
\end{align}
%%%
Since  we have 
%%%
\begin{align}
\int \!d\bz\; \qphi(\bz) \nabla_\bphi \log \qphi(\bz) = \nabla_\bphi \int \!d\bz\; \qphi(\bz) =  \nabla_\bphi1 = 0
\end{align}
%%%
we can simplify \eqnref{eqn:phigradsupp2} further as 
%%%
\begin{align}
\label{eqn:phigradsupp3}
    \nabla_\bphi \elbo \rightarrow \EE_{\qphi(\bz)}\left[ \left(\log \pth(\bx, \bz) - \log \qphi(\bz) \right) \nabla_\bphi \log \qphi(\bz) \right]
\end{align}
%%%
which is precisely the conventional score function (i.e.~REINFORCE-like) gradient estimator for the ELBO, used e.g.~in \citep{ranganath2014black}.
%%%%%5
The VRS gradient estimator for model parameters $\bth$ can also be expressed in a number of different ways:
%%%
\begin{align}
\label{eqn:thetagradsupp}
\nabla_\bth \elbo &= \EE_{\rr(\bz)} \left[ \nabla_\bth \log \pth(\bx, \bz) \right] +
 {\rm COV}_{\rr(\bz)} \left[\AAA(\bz), \nabla_\bth \log \aaa(\bz) \right] \\ \nonumber
&= \EE_{\rr(\bz)} \left[ \nabla_\bth \log \pth(\bx, \bz) \right] -
 {\rm COV}_{\rr(\bz)} \left[\AAA(\bz), \sigma(\ellT(\bz)) \nabla_\bth \log \pth(\bx, \bz) \right] \\ \nonumber
 &= \EE_{\rr(\bz)} \left[ \nabla_\bth \log \pth(\bx, \bz) \right] -
 {\rm COV}_{\rr(\bz)} \left[\AAA(\bz), (1 - \aaa(\bz)) \nabla_\bth \log \pth(\bx, \bz) \right]
\end{align}
%%%
In the limit that $T\rightarrow\infty$ we have 
%%%
\begin{align}
\nabla_\bth \elbo \rightarrow \EE_{\qphi(\bz)} \left[ \nabla_\bth \log \pth(\bx, \bz) \right] 
\end{align}
%%%
which, as we would expect, is the conventional ELBO gradient estimator for model parameters.

%%%%%%%%%%%%%%%%%%%%%%%%%%%%%%%%%
\subsection{Monte Carlo Estimation}
\label{sec:vrsmc}
%%%%%%%%%%%%%%%%%%%%%%%%%%%%%%%%%

Due to the covariance terms obtaining unbiased Monte Carlo estimates of the 
gradient estimators \eqnref{eqn:phigradsupp} and \eqnref{eqn:thetagradsupp} 
requires drawing $S>1$ samples from $\rr(\bz)$. To do so we appeal to the identity in \eqnref{eqn:covidentity}.
For example we can approximate the $\bphi$ gradient estimator as follows:
%%%
\begin{align}
    \nabla_\bphi \elbo &= {\rm COV}_{\rr(\bz)} \left[ \AAA(\bz), \aaa(\bz) \nabla_\bphi \log \qphi(\bz) \right] \\
    &\approx
    \tfrac{1}{S-1}\sum_{s=1}^S \left\{  \AAA(\bz_s) - \tfrac{1}{S}\Sigma_{s^\prime} \AAA(\bz_{s^\prime})\right\} 
    \aaa(\bz_s) \nabla_\bphi \log \qphi(\bz_s)
\end{align}
%%%

%%%%%%%%%%%%%%%%%%%%%%%%%%%%%%%%%%%%%
\section{Adaptively tuning $T$}
\label{sec:apptuning}
%%%%%%%%%%%%%%%%%%%%%%%%%%%%%%%%%%%%%

As detailed in \secref{sec:tuning} we can adjust the rejection threshold
$T$ using the gradient
%%%
\begin{align}
\frac{\partial \LL}{\partial T} = \left(\ZZ_r \! - \! \ZZ_\target \right) \EE_{\qphi(\bz)}\!\left[ \frac{\partial \aaa(\bz)} {\partial T} \right]
=  \EE_{\qphi(\bz)}\left[ \aaa(\bz) \! - \! \ZZ_\target \right] \EE_{\qphi(\bz)}\left[ \aaa(\bz) (1 \!-\! \aaa(\bz))\right] \nonumber
 \end{align}
%%%
To obtain an unbiased Monte Carlo estimate of this quantity we draw $S>1$ samples from $\rr(\bz)$
and use the same logic used to derive \eqnref{eqn:covidentity} to compute 
%%%
\begin{align}
\frac{\partial \LL}{\partial T} \approx
    \widehat{\frac{\partial \LL}{\partial T}} = 
    \tfrac{1}{S} \sum_{s=1}^S \left\{ \aaa(\bz_{s})(1-\aaa(\bz_{s}))\left(\tfrac{1}{S-1}(\Sigma_{s^\prime=1}^S\aaa(\bz_{s^\prime})-\aaa(\bz_{s})) - \ZZ_\target\right)\right\}
 \end{align}
%%%
While this stochastic gradient estimator could be plugged into a variety of optimization algorithms, 
for simplicity we use vanilla SGD (stochastic gradient descent) with a fixed learning rate of $1$.
In other words at each step $t$ in RVRS ELBO optimization we make the update
%%%
\begin{align}
    T_{t+1} = T_t - \widehat{\frac{\partial \LL}{\partial T} }
 \end{align}
%%%
We find that this works well in practice---in particular on all the experiments reported here---although we expect that 
more sophisticated schemes could perform better. We also note that perfect adaptation of $T$ is not necessary, since---provided
$T$ is in the right ballpark---the primary relevance of $T$ is to determine the precise computation to inference fidelity trade-off. For example if we set $\ZZ_\target=0.30$ but end up with $\ZZ_r = 0.29$ the result is that we used a bit more computation then we intended---and obtained a slightly better variational approximation as a result.

%%%%%%%%%%%%%%%%%%%%%%%%%%%%%%%%%%%%%
\section{Semi-RVRS}
\label{sec:semisupp}
%%%%%%%%%%%%%%%%%%%%%%%%%%%%%%%%%%%%%

The variational distribution for Semi-RVRS is given by
%%%%%
\begin{align}
    \label{eqn:semirvrspost}
    \qphi(\bz_G) \prod_{n=1}^N \rrn(\bz_n|\bz_G) = \frac{1}{\ZZ_r} \qphi(\bz_G) \prod_{n=1}^N \qphin(\bz_n) \aaan(\bz_n | \bz_G)
\end{align}
%%%%%
where we assume for simplicity that $\qphin(\bz_n)$ does not depend explicitly on $\bz_G$ (though this could
easily be accommodated). Here $\qphi(\bz_G)$ is some reparameterizable and parametric variational distribution and 
each distribution $\rrn(\bz_n|\bz_G)$ is given by
%%%
\begin{equation}
\rrn(\bz_n|\bz_G) \propto \qphin(\bz_n) \aaan(\bz_n | \bz_G)
\end{equation}
%%%
with
%%%
\begin{equation}
\aaan(\bz_n | \bz_G) \equiv \sigma(\log \pth(\bx_n | \bz_n, \bz_G) \pth(\bz_n | \bz_G) - \log \qphin(\bz_n) + T_n)
\end{equation}
%%%
and where each each $T_n \in \RR$ is a rejection threshold parameter.
For details on sampling from \eqref{eqn:semirvrspost} and ELBO computation see the next section, \secref{sec:semirvrselbo}. 
For details on estimating the normalization constant $\ZZ_r$ for the purposes of evaluation see \secref{sec:semirvrspartition}.

%%%%%%%%%%%%%%%%%%%%%%%%%%%%%%%%%%%%%%%%%%%%%%%%%%%%%%%%%%%%%%%%%%%%%%%
\subsection{ELBO computation and rejection sampling on a parallel machine}
\label{sec:semirvrselbo}
%%%%%%%%%%%%%%%%%%%%%%%%%%%%%%%%%%%%%%%%%%%%%%%%%%%%%%%%%%%%%%%%%%%%%%%

The ELBO for Semi-RVRS is given by
%%%%%
\begin{align}
    \label{eqn:semirvrselbo}
    \EE_{\qphi(\bz_G) \prod_n \rrn(\bz_n|\bz_G)} \Big[
        &\log \pth(\bz_G) + \Sigma_n \log\{\pth(\bx_n | \bz_n, \bz_G) p(\bz_n | \bz_G)\} \nonumber \\
        &-\log \qphi(\bz_G) - \Sigma_n \log \rrn(\bz_n|\bz_G) \Big]
\end{align}
%%%%%
To construct Monte Carlo gradient estimates of \eqref{eqn:semirvrselbo} we proceed as follows.
First we randomly choose a mini-batch of data of size $B$ specified by unique indices $\{i_1,\ldots,i_B\}$
and draw a sample of the global latent variable $\bz_G \sim \qphi(\bz_G)$.
Next we either run the (potentially slow) unbiased sampler defined in Algorithm~\ref{alg:semiunbiased};
otherwise we run the (potentially much faster) biased sampler defined in Algorithm~\ref{alg:semibiased}.
In Algorithm~\ref{alg:semiunbiased} we always return exactly $S$ samples $\bz_n^{1:S} $ for each
data point $n$. Since a variable number of proposals may need to be drawn for each data point before
this is the case, the runtime of this algorithm can be pretty variable (although this variability can be mitigated
by dynamically reallocating compute resources, see Algorithm~\ref{alg:semiunbiased}).
Since however we have exactly $S$ samples for each data point it is straightforward to follow
the recipe in \secref{sec:phigrad} to construct an unbiased gradient estimator of the Semi-RVRS ELBO \eqref{eqn:semirvrselbo}. If instead we use Algorithm~\ref{alg:semibiased} some data points in the mini-batch may have fewer
than $S$ accepted samples. 
Consequently we do not use these data points in constructing our Monte Carlo ELBO gradient estimators
(note that we need to appropriately re-scale terms in our Monte Carlo estimator to account for the effectively
variable mini-batch size). This introduces some bias, however it makes our Semi-RVRS ELBO gradient
estimators quite a bit faster (especially for small $\ZZ_\target$), since we do not need to waste compute
on `stragglers', i.e.~data points that have fewer than $S$ accepted samples. Note that the resulting
bias is not expected to be too severe, since the bias is exactly zero if the local acceptance probabilities
of each data point are equal (e.g.~if they are all exactly equal to $\ZZ_\target$). While this condition
never holds exactly, it holds approximately if the adaptation of the $\{T_n\}$ is working well, and
this is enough to ensure that the bias is minimal provided that $S^\prime$ in Algorithm~\ref{alg:semibiased} is
sufficiently large so that most data points in each mini-batch (say $>80-90\%$) are accepted. As a rule of thumb
one might choose $S^\prime = {\rm ceil}(S / \ZZ_\target)$ or $S^\prime = {\rm ceil}(2S / \ZZ_\target)$.
See \figref{fig:biasedsemi} in \secref{sec:addexp} for empirical confirmation of this intuition.

Note that the above discussion has focused on the more general case of Semi-RVRS with both global and local
latent variables. However the basic logic of Algorithm~\ref{alg:semiunbiased} and Algorithm~\ref{alg:semibiased}
is also applicable in the case with purely local latent variables: just ignore the global latent variable. 
Indeed we use Algorithm~\ref{alg:semibiased}
when training VAEs in \secref{sec:vae} and Algorithm~\ref{alg:semiunbiased} when evaluating VAE ELBOs after training.

%%%%%%%%%%%%%%%%%%%%%%%%%%%%%
\begin{algorithm}[t]
	\caption{Unbiased sampler for the Semi-RVRS variational distribution in \eqref{eqn:semirvrspost}. 
    The same algorithm can also be used for the case with only local latent variables. 
    Optionally dynamically reallocate compute resources to focus on data points that do not have $S$ accepted samples.
    {\bf Input}: subsample indices $\{i_1,\ldots,i_B\}$, number of samples $S$ per data point, acc.~prob.~$\{a_{\bphi_n, \bth}(\bz_n)\}$, and proposals $\{q_{\bphi_n}(\bz_n)\}$.
		%%{\bf Output}: $S$ sets of samples $\{\bz_{i_1}^{1:S},\ldots,\bz_{i_M}^{1:S}\}$.
	}
	\label{alg:semiunbiased}
	\begin{algorithmic}[1]
       \For{$k \gets 1$ to $B$} \Comment{Initialize the number of accepted samples for each data point}
       	\State $s_k \gets 0$
       \EndFor
		\While{$\min\{s_1,\ldots,s_B\} < S$}
            \If {dynamically reallocating compute}
		    \For{$k \gets 1$ to $B$} \Comment{Compute how many samples are left to draw}
		    	\State $w_k \gets \max\{S - s_k, 0\}$
		    \EndFor
            \EndIf
		    \For{$k \gets 1$ to $B$}
                \If {dynamically reallocating compute}
                    \State         $j \sim {\rm Categorical}(\frac{w_1}{\sum_m w_m},\ldots,\frac{w_B}{\sum_m w_m})$
                \Else
                    \State  $j \gets k$
				\EndIf
			    \State $n \gets i_j$
                \State $\bz_n \sim q_{\bphi_n}(\bz_n)$ \Comment{Draw from proposal distribution}
                \If {$u < a_{\bphi_n, \bth}(\bz_n)$ where $u \sim {\rm Uniform}(0,1)$} \Comment{Do rejection sampling}
                    \State $s_j \gets s_j + 1$ \Comment{Keep track of number of accepted samples for each data point}
					\State $\bz_{n}^{s_j} \gets \bz_n$
				\EndIf
		    \EndFor
		\EndWhile
        \State \textbf{return} $\{\bz_{i_1}^{1:S},\ldots,\bz_{i_B}^{1:S}\}$ \Comment{Return exactly $S$ samples for each data point}
	\end{algorithmic}
\end{algorithm}
%%%%%%%%%%%%%%%%%%%%%%%%%%%%%

%%%%%%%%%%%%%%%%%%%%%%%%%%%%%
\begin{algorithm}[t]
	\caption{Biased sampler for the Semi-RVRS variational distribution in \eqref{eqn:semirvrspost}. {\bf Input}: subsample indices $\{i_1,\ldots,i_B\}$, number of samples $S$ per data point, number of candidates $S'\ge S$, acc.~prob.~$\{a_{\bphi_n, \bth}(\bz_n)\}$, and proposals $\{q_{\bphi_n}(\bz_n)\}$.
    The same algorithm can also be used for the case with only local latent variables. 
		%%{\bf Output}: $S$ sets of samples $\{\bz_{i_1}^{1:S},\ldots,\bz_{i_M}^{1:S}\}$.
	}
	\label{alg:semibiased}
	\begin{algorithmic}[1]
	  \For{$k \gets 1$ to $B$}
	    \For{$t \gets 1$ to $S'$}
	      \State $\bz_{i_k}^t \sim q_{\phi_{i_k}}(\bz_{i_k})$
	      \State $u \sim {\rm Uniform}(0,1)$
		  \State $\mathrm{acc}_k^t  \gets u < a_{\bphi_{i_k}, \bth}(\bz_{i_k})$
	    \EndFor
        \State $\mathrm{j}_{1:S'} \gets \mathrm{argsort}(\mathrm{acc}_k^{1:S'})$ \Comment{Acc.~samples thus have larger indices than non-acc.~samples}
	    \For{$t \gets 1$ to $S'$}
	      \State $\bz_{i_k}^t \gets \bz_{i_k}^{j_{S-t + 1}}$
	      \State $\mathrm{acc}_{k}^t \gets \mathrm{acc}_{k}^{j_{S-t + 1}}$
	    \EndFor
        \State $\mathrm{mask}_k \gets (\sum_{t=1}^S \mathrm{acc}_k^t = S) $
	  \EndFor
        \State \textbf{return} $\{(\bz_{i_1}^{1:S}, \mathrm{mask}_1),\ldots,(\bz_{i_B}^{1:S}, \mathrm{mask}_B)\}$ \Comment{Return mask and $S$ samples for each data point}
	\end{algorithmic}
\end{algorithm}
%%%%%%%%%%%%%%%%%%%%%%%%%%%%%

%%%%%%%%%%%%%%%%%%%%%%%%%%%%%%%%%%%%%%%%%%%%%%%%%%%%%%%%%%%%%%%%%%%%%%%
\subsection{The Semi-RVRS normalization constant}
\label{sec:semirvrspartition}
%%%%%%%%%%%%%%%%%%%%%%%%%%%%%%%%%%%%%%%%%%%%%%%%%%%%%%%%%%%%%%%%%%%%%%%

The normalization constant $\ZZ_r$ for the Semi-RVRS variational distribution in \eqref{eqn:semirvrspost} is given by 
%%%%%
\begin{align}
\ZZ_r \equiv \EE_{\qphi(\bz_G)} \prod_{n=1}^N \EE_{\qphin}(\bz_n) \left[ \aaan(\bz_n | \bz_G) \right]
\label{eqn:localZ}
\end{align}
%%%%%
To compute the corresponding ELBO for evaluation purposes we need to estimate
the quantity $\log \ZZ_r$, since the ELBO is given by
%%%%%
\begin{align}
    {\rm ELBO} &= \EE_{\qphi(\bz_G)} \EE_{\rr(\bz_{1:N}|\bz_G)} \left[ 
    \log \pth(\bz_G, \bz_{1:N}) - \log \rr(\bz_{1:N}|\bz_G) \right]  \\
    &= \EE_{\qphi(\bz_G)} \EE_{\rr(\bz_{1:N}|\bz_G)} \left[
    \log \pth(\bz_G, \bz_{1:N}) - \log \qphi(\bz_{1:N}) -\log \aaa(\bz_{1:N} | \bz_G) + \log \ZZ_r \right] \nonumber
\end{align}
%%%%%
where we for convenience we write 
%%%%%
\begin{align}
    &\rr(\bz_{1:N} | \bz_G) = \prod_{n=1}^N \rrn(\bz_n|\bz_G) \qquad 
    \qphi(\bz_{1:N}) = \prod_{n=1}^N \qphin(\bz_n) \nonumber \\
    &\aaa(\bz_{1:N} | \bz_G) = \prod_{n=1}^N \aaan(\bz_n | \bz_G) 
\end{align}
%%%%%
Unfortunately it is difficult to construct an unbiased low variance
estimator for $\log \ZZ_r$. Indeed, although the naive plug-in Monte Carlo estimator for \eqnref{eqn:localZ}
is consistent, it is biased and is generally expected to be high variance. Consequently
\emph{for the purposes of evaluation only}\footnote{Recall that the ELBO gradient estimators we use,
which are based on Prop.~\ref{prop:gradest}, are unbiased and low variance.} we replace $\log \ZZ_r$ with a lower bound that is easier to estimate. 
Indeed we just appeal to Jensen's inequality to obtain
%%%%%
\begin{align}
\log \ZZ_r &\equiv \log \EE_{\qphi(\bz_G)} \prod_{n=1}^N \EE_{\qphin}(\bz_n) \left[ \aaan(\bz_n | \bz_G) \right] \\
 &\ge \EE_{\qphi(\bz_G)} \log \prod_{n=1}^N \EE_{\qphin}(\bz_n) \left[ \aaan(\bz_n | \bz_G) \right] \\
    &\equiv \LL^{{\rm lb}} = \EE_{\qphi(\bz_G)} \sum_{n=1}^N \log \EE_{\qphin}(\bz_n) \left[ \aaan(\bz_n | \bz_G) \right]
\label{eqn:localZbound}
\end{align}
%%%%%
While the plug-in Monte Carlo estimator for $\LL^{{\rm lb}}$ in \eqnref{eqn:localZbound} is still biased because
the expectations w.r.t.~$\bz_n$ occur inside of a logarithm, the important point is that
$\LL^{{\rm lb}}$ is consistent and low variance. Indeed for local latent variables that are relatively low-dimensional,
the plug-in Monte Carlo estimator for $\EE_{\qphin}(\bz_n) \left[ \aaan(\bz_n | \bz_G) \right]$ is expected to be low-variance
and so the bias will be correspondingly small. As such the use of $\LL^{{\rm lb}}$ in evaluating Semi-RVRS
ELBOs is expected to yield high-fidelity low-variance approximations to the exact ELBO, and it is these
estimators that we report in our experiment in \secref{sec:local}. To be precise we use the following nested Monte Carlo estimator
%%%%%
\begin{align}
    \log &\ZZ_r \approx \frac{1}{M_1} \sum_{m_1=1}^{M_1} \sum_{n=1}^N  \log \left\{ \frac{1}{M_2} \sum_{m_2=1}^{M_2} \aaan(\bz_{n,m_1,m_2} | \bz_{G,m_1})\right\} \\
    &{\rm with} \quad \bz_{G,m_1} \sim \qphi(\cdot) \qquad {\rm and} \qquad \bz_{n,m_1,m_2} \sim \qphin(\cdot)\nonumber
    \\&{\rm for} \;\; n=1,...,N \;\; {\rm and}  \;\; m_1=1,...,M_1 \;\; {\rm and}  \;\; m_2=1,...,M_2 \nonumber
\end{align}
%%%%%
with $M_1=10^4$ and $M_2=10^3$.

%%%%%%%%%%%%%%%%%%%%%%%%%%%%%%%%%%%%%
\section{Experimental details}
\label{sec:expdetails}
%%%%%%%%%%%%%%%%%%%%%%%%%%%%%%%%%%%%%

%%%%%%%%%%%%%%%%%%%%%%%%%%%%%%%%
\subsection{General RVRS details}
%%%%%%%%%%%%%%%%%%%%%%%%%%%%%%%%

We always use $S=2$ samples to compute multi-sample RVRS ELBO gradient estimators during training.
In all cases we use either mean-field or multivariate\footnote{With Cholesky-parameterized full-rank covariance matrices.} Normal
proposal distributions $\qphi(\bz)$.
Similar to \citep{geffner2021mcmc} in the context of UHA, we initialize the RVRS proposal distribution with a variational distribution
obtained by maximizing a conventional ELBO. 
We initialize the rejection threshold $T$ to minus the ELBO obtained with mean-field variational inference.
We use the Adam optimization algorithm for all ELBO optimization \cite{kingma2014adam}.
For RVRS we use an initial learning rate of $10^{-4}$ that is is decimated twice over the course of training: after $1/3$ and $2/3$
of the total number of training iterations. 
Unless specified otherwise we used $\epsilon=10^{-4}$ (see \eqnref{eqn:aaaeps}).

%%%%%%%%%%%%%%%%%%%%%%%%%%%%%%%%
\subsection{Other experimental details}
%%%%%%%%%%%%%%%%%%%%%%%%%%%%%%%%

Like RVRS we initialize UHA base distributions with a variational distribution obtained by maximizing a conventional ELBO.
For UHA we use an initial learning rate of $10^{-4}$ that is is decimated twice over the course of training: after $1/3$ and $2/3$ of the total number of training iterations.
For UHA we limit the stepsize $\eta$ to $\eta_{\rm max} = 0.25$ and initialize step sizes to $\eta = 0.005$.
UHA ELBOs are computed using a single sample Monte Carlo estimate during training.
For mean-field, IWAE, and flow training we use an initial learning rate of $10^{-3}$ that is is decimated twice over the course of training: after $1/3$ and $2/3$ of the total number of training iterations.
Mean-field and normalizing flow ELBOs are computed using a single sample Monte Carlo estimate during training.
Just like for RVRS we use the Adam optimization algorithm for all variational baselines \cite{kingma2014adam}.
For the Block Neural Autoregressive normalizing flow \citep{de2020block}
we use \texttt{AutoBNAFNormal} implemented in NumPyro with default settings (in particular one layer).

%%%%%%%%%%%%%%%%%%%%%%%%%%%%%%%%
\subsection{Datasets}
%%%%%%%%%%%%%%%%%%%%%%%%%%%%%%%%

Apart from MNIST we use a number of UCI \citep{asuncion2007uci} datasets: 
MiniBooNE, SUSY, Higgs, Adult, Bank, Mushroom, Thyroid, Spambase, Pol, \& Bike.

%%%%%%%%%%%%%%%%%%%%%%%%%%%%%%%%
\subsection{Characterizing RVRS}
%%%%%%%%%%%%%%%%%%%%%%%%%%%%%%%%

The log density of the non-gaussian target in \figref{fig:funnel} is given by the formula
%%%%%
\begin{align}
    \log \phi(\tfrac{x + y}{\sqrt{2}} | 0, 1) + \log \phi(\tfrac{x - y}{\sqrt{2}} | 0, e^{\tfrac{x + y}{\sqrt{2}}})
\end{align}
%%%%%
where $\phi(x|\mu,\sigma^2)$ denotes the density of a Normal distribution with mean $\mu$ and variance
$\sigma^2$ evaluated at $x$. To train variational approximations we train for $5$ million gradient steps.
We evaluate ELBOs with $1$ million samples and use $\epsilon=10^{-6}$ for RVRS.

The gradient variance results depicted in \figref{fig:gradvar} were obtained as follows. We
use $N=100$ data points from the MiniBooNE UCI dataset, which has $D=51$ covariate dimensions. Additional covariate
dimensions are removed (via subsetting the original covariates) or added as needed by sampling i.i.d.~from a standard normal distribution.
Both VRS and RVRS mean-field gaussian proposal distributions are initialized by optimizing a conventional ELBO
for $1000$ steps. The threshold $T$ is set to minus the ELBO. Variance estimates are made with $5 \times 10^5$ samples.

The results in \figref{fig:varyZ} were also obtained using $N=100$ data points from the MiniBooNE UCI dataset.
We train for $2.4$ million steps and consider $\ZZ_\target$ ranging from $0.004$ to $0.40$.
See \figref{fig:fitZ} for additional results pertaining to this experiment.

%%%%%%%%%%%%%%%%%%%%%%%%%%%%%%%%
\subsection{Logistic regression}
%%%%%%%%%%%%%%%%%%%%%%%%%%%%%%%%

We do a total of $3 \times 10^5$ training iterations for the normalizing flow due to its computational cost.
For all other methods we do a total of $9 \times 10^5$ training iterations.
The datasets we use were subsampled down to $N=100$ training data points.
This choice was made to ensure a non-trivial amount of non-gaussianity and to enable a comparison with HMC.
%$10^4$ data points were used for testing.  
$10^5$ samples were used for ELBO evaluation for all methods. Timing results are reported
using a machine with an AMD EPYC 7R13 CPU.

We used NUTS implemented in NumPyro to generate the samples used to compute Max Slice Wasserstein distances.
We used a diagonal mass matrix and $10^4$ warmup steps. We generated $5 \times 10^5$ post-warmup samples. 
Every $5^{\rm th}$ sample was retained for a total of $10^5$ samples. We then drew $10^5$ independent samples
from each variational method. These samples were then used to compute Max Slice Wasserstein distances
using POT \citep{flamary2021pot}.
To compute each Wasserstein distance we use $1000$ random projections and average results across $10$ replicates.

%%%%%%%%%%%%%%%%%%%%%%%%%%%%%%%%
\subsection{Gaussian process classification}
%%%%%%%%%%%%%%%%%%%%%%%%%%%%%%%%

We used $N=256$ data points for training for each dataset. We used a RBF kernel with per-dimension lengthscales and a logistic
link function with a Bernoulli likelihood. We trained for $6 \times 10^5$ iterations and used $2 \times 10^4$ 
samples for ELBO evaluation. For all methods the base/proposal distribution used is a multivariate Normal
distribution with a Cholesky-parameterized full-rank covariance matrix.
Due to the delicate linear algebra we do all computations in 64-bit precision.

%%%%%%%%%%%%%%%%%%%%%%%%%%%%%%%%
\subsection{Variational autoencoders}
%%%%%%%%%%%%%%%%%%%%%%%%%%%%%%%%

For all methods we used the same batch size ($B=100$), trained for $1500$ epochs, 
and evaluated using $5000$ samples.  The training/test set consist of $60$k/$10$k images, respectively.
The latent variable has a standard Gaussian prior and dimension $D=50$.
Both the encoder and decoder are multilayer perceptrons with two hidden layers of $200$ hidden units and with
tanh activation functions.
All experiments were done on a RTX 2070 GPU with 8GB of memory.
We used the Adam optimizer and learning rates were decimated, i.e.~reduced by a factor of $10$, at $500$ and $1000$ epochs.
When training with a conventional ELBO, IWAE, UHA, and RVRS the initial learning rates were $10^{-3}$, $10^{-3}$, $10^{-4}$, and $10^{-4}$, respectively.
For both UHA and RVRS encoder-decoder parameters were initialized using the final optimized parameters obtained after training with a conventional ELBO. 
In UHA we used the same set of (learned) step sizes and mass matrices for all data points,~i.e.~only the
base distribution is amortized. In RVRS we used the biased sampler Algorithm \ref{alg:semibiased} with $S=2$ and 
$S' = {\rm round}(S / \ZZ_\target)$ for training. 
Evaluation was done with Algorithm \ref{alg:semiunbiased}.

We initialize the threshold parameter $T_n$ in RVRS for each training data point to a $50$-sample Monte Carlo estimate of its corresponding negative ELBO (obtained with the mean field proposal $\qphi(\bz)$). Since we do not amortize $T_n$, after training we need to choose $T_n$ for each unseen test data point such that the acceptance probability of the rejection sampler will approximately equal $\ZZ_\target$. 
Hence for each test data point, we draw $50$ samples $\{\bz_n\}$ from the proposal distribution $\qphi$ and choose $\{T_n\}$ 
to minimize the objective $\LL(T_n) = \tfrac{1}{2} \left(\ZZ_{r,n} - \ZZ_\target \right)^2$ for each data point.

%%%%%%%%%%%%%%%%%%%%%%%%%%%%%%%%
\subsection{Hierarchical modeling}
%%%%%%%%%%%%%%%%%%%%%%%%%%%%%%%%

Both datasets we use have $5000$ data points. We add additional Normally distributed noise to $25\%$ of the
data points to drive the model into a regime where the Student's t likelihood is needed to model the resulting
heavy-tailed noise.
We use $6 \times 10^5$ training iterations for all methods and a mini-batch size of $256$.
Due to the special functions involved in the Gamma probability density function we do all computations in 64-bit precision.

%%%%%%%%%%%%%%%%%%%%%%%%%%%%%%%%%%%%%
\section{Additional experimental results}
\label{sec:addexp}
%%%%%%%%%%%%%%%%%%%%%%%%%%%%%%%%%%%%%

\begin{figure}[ht]
\begin{center}
\includegraphics[width=0.5\textwidth]{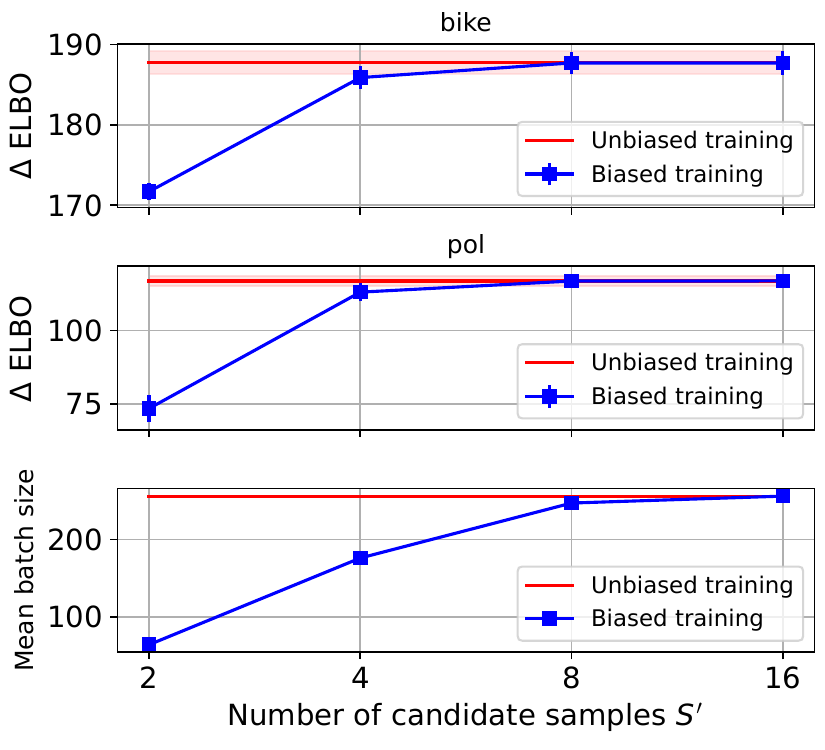}
\caption{We compare Semi-RVRS training for the hierarchical model in \secref{sec:local}
    using Algorithm~\ref{alg:semibiased} (in blue) and Algorithm~\ref{alg:semiunbiased} (in red).
    We consider the same two datasets: bike and pol.
    Uncertainty bands/bars denote 90\% confidence intervals obtained from $5$ independent runs.
    As expected provided $S^\prime$ is sufficiently large so that the mean batch size is a large
    fraction of $B=256$, then the bias introduced by `dropping stragglers' 
    is minimal and the performance of Algorithm~\ref{alg:semibiased} 
    approaches that of Algorithm~\ref{alg:semiunbiased}. 
    See \secref{sec:semisupp} for additional discussion.
    }
    \label{fig:biasedsemi}
\end{center}
\end{figure}

\begin{figure}[ht]
\begin{center}
\includegraphics[width=0.6\textwidth]{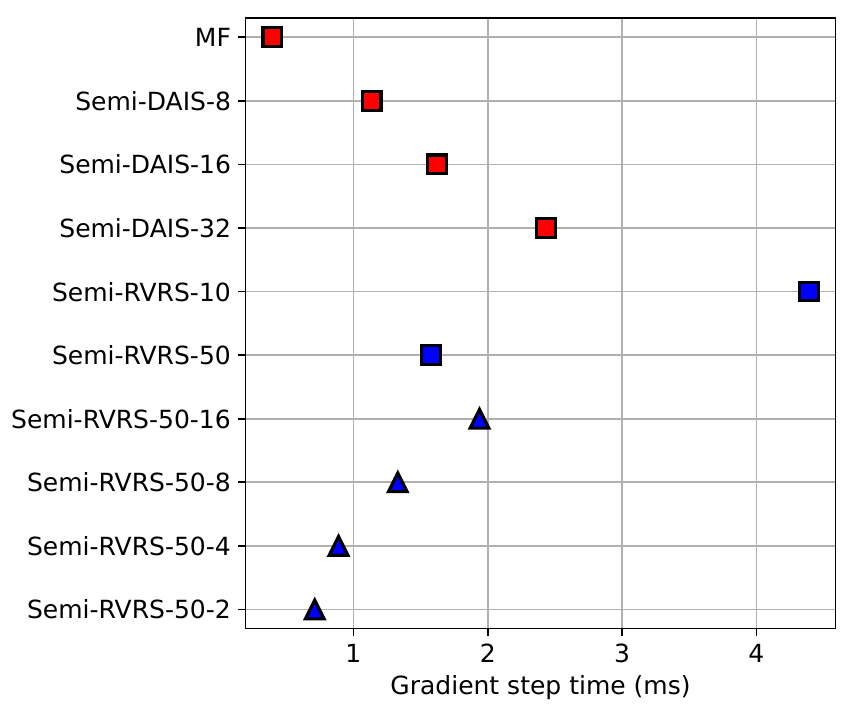}
\caption{We report training times for the hierarchical model in \secref{sec:local}
    on the bike dataset. We compare baseline methods (red) to RVRS variants (blue).
    Among RVRS variants we compare methods using Algorithm~\ref{alg:semiunbiased} (squares)
    to methods using Algorithm~\ref{alg:semibiased} (triangles).
    Notably Semi-RVRS-$0.50$-$8$, i.e.~Semi-RVRS with $\ZZ_\target=0.50$ and
    $S^\prime=8$, significantly outperforms e.g.~Semi-DAIS-$8$ (see Table~\ref{table:local}
    and \figref{fig:biasedsemi}) but is faster.
    Timing results are obtained using a machine with an AMD EPYC 7R13 CPU and
    make it clear that Algorithm~\ref{alg:semibiased} can be significantly faster than Algorithm~\ref{alg:semiunbiased}
    if $S^\prime$ is moderate.
    }
    \label{fig:biasedsemitimes}
\end{center}
\end{figure}

\begin{figure}[ht]
\begin{center}
\includegraphics[width=0.8\textwidth]{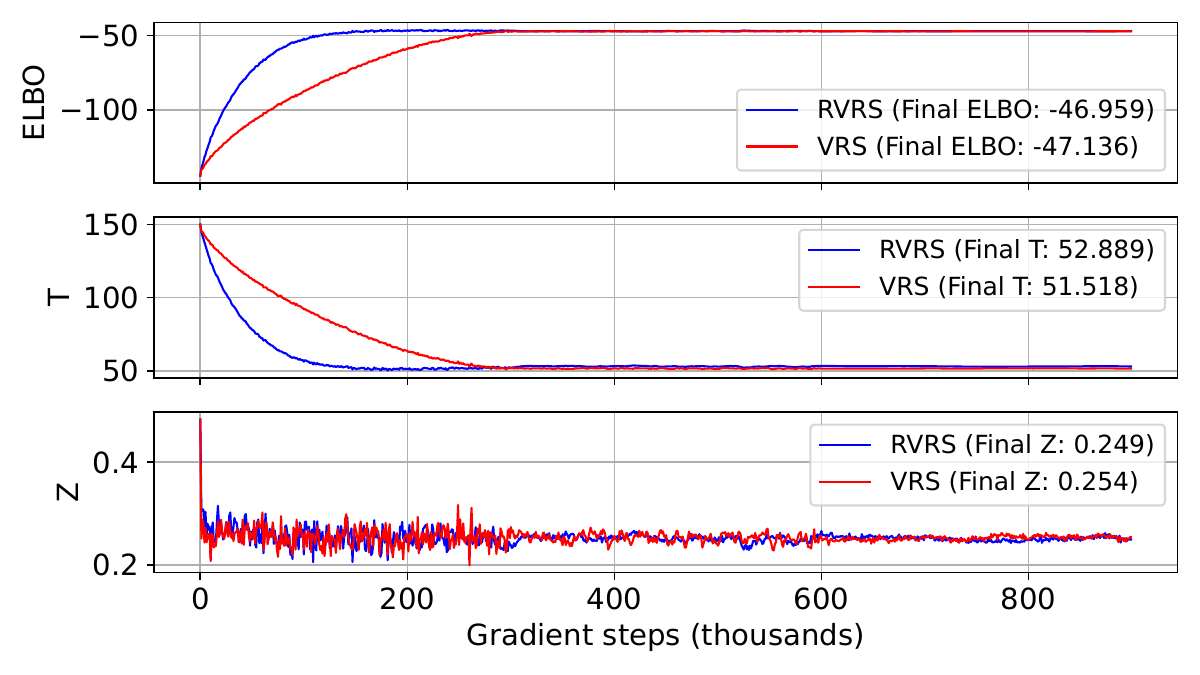}
\caption{We compare RVRS and VRS training curves for a logistic regression problem with $N=100$ data points
    and a $D=51$ dimensional latent space for $\ZZ_\target=0.25$. From top to bottom we depict the ELBO (computed with 20k samples
    every 1000 steps), the threshold parameter $T$, and the value of $\ZZr$ (computed with 20k samples
    every 1000 steps). The initial learning rate is $10^{-4}$ and is decimated at 300k and 600k steps.
    Due to the lower gradient variance of RVRS, RVRS ELBO training makes more rapid progress. 
    For example RVRS attains an ELBO of $-50$ after $\sim$115k steps, while VRS does not attain this value until $\sim$255k steps.
    Similarly RVRS attains an ELBO of $-100$ after $\sim$24k steps, while VRS does not attain this value until $\sim$69k steps.
    }
    \label{fig:elbocurves}
\end{center}
\end{figure}

\begin{figure}[ht]
\begin{center}
\includegraphics[width=0.49\textwidth]{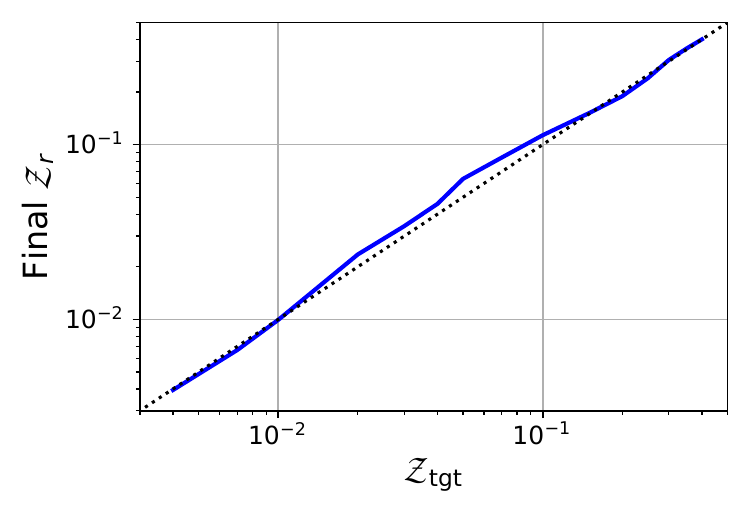}
\includegraphics[width=0.49\textwidth]{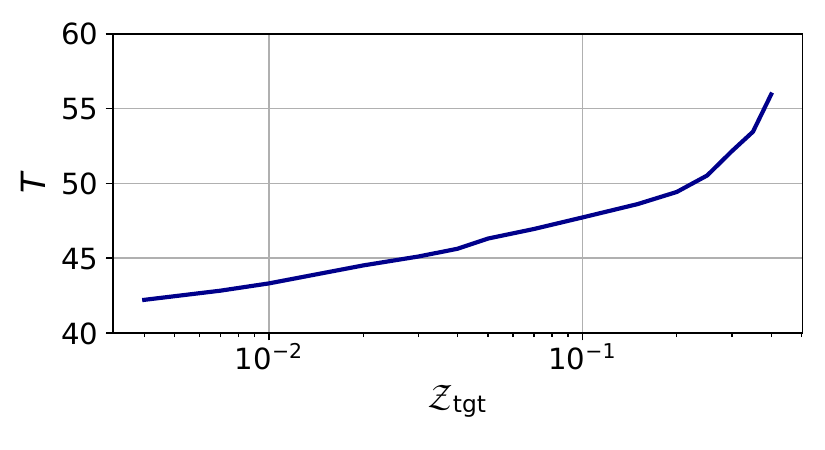}
    \caption{
    We explore the performance of RVRS as a function of $\ZZ_\target$ on a logistic
    regression problem in $D=51$ dimensions. {\bf (Left)} We show that the $T$ adaptation scheme 
    described in \secref{sec:tuning} and \secref{sec:apptuning} works well over a broad range of $\ZZ_\target$.
    {\bf (Right)} We show how the adapted $T$ changes as a function of $\ZZ_\target$.
    Note that this is a companion figure to \figref{fig:varyZ}.
    }
    \label{fig:fitZ}
\end{center}
\end{figure}

\begin{figure}[ht]
\begin{center}
\includegraphics[width=0.9\textwidth]{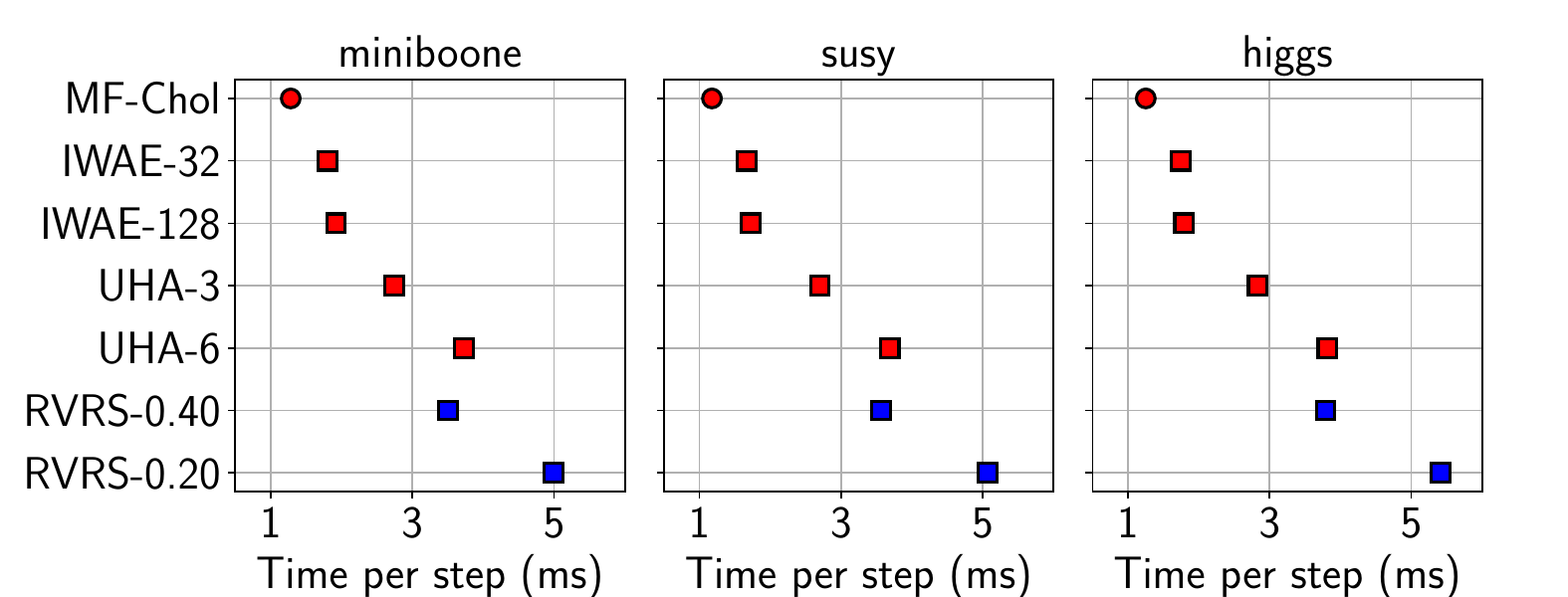}
    \caption{
    In this companion figure to \figref{fig:gp} we report times per gradient step for
    the GP classification experiment in \secref{sec:gp}. Results are obtained with
    a NVIDIA Tesla V100 GPU. Note that the relative speed of IWAE is a quirk of this particular regime.
    For relatively moderately sized matrices (here $256 \times 256$) commercial GPUs like the V100 can
    compute a large number of Cholesky decompositions in parallel. As such IWAE parallelizes particularly
    well in this regime. For larger matrices (e.g.~$1024 \times 1024$) this advantage would largely
    evaporate. We also note that we could adapt Algorithm~\ref{alg:semibiased} to the setting without
    local latent variables, which could make RVRS significantly faster when running on parallel-friendly
    hardware like a GPU.
    }
    \label{fig:gpgrad}
\end{center}
\end{figure}

In \figref{fig:biasedsemi} and \figref{fig:biasedsemitimes} we report additional results pertaining to the experiment
in \secref{sec:local}.
In \figref{fig:elbocurves} we compare the training dynamics of VRS and RVRS.
In \figref{fig:fitZ} we explore the performance of our $T$ adaptation scheme.
In \figref{fig:gpgrad} we report times per gradient step for the GP experiment in \secref{sec:gp}.
In Table \ref{table:vaebig} we report additional results for the VAE experiment in \secref{sec:vae}.

%%%%%%%%%%%%%%%%%%%%
%%%%%%%%%%%%%%%%%%%%

%%%%%%%%%%%%%%%%%%%%
%%%%%%%%%%%%%%%%%%%%

%%%%%%%%%%%%%%%%%%%%
%%%%%%%%%%%%%%%%%%%%

%%%%%%%%%%%%%%%%%%%%
%%%%%%%%%%%%%%%%%%%%

%%%%%%%%%%%%%%%%%%%%
%%%%%%%%%%%%%%%%%%%%

\begin{table}[H]
	\centering
	\resizebox{.99\textwidth}{!}{%
		\begin{tabular}{l||l|l|l|l|l|l|l|l|l}
			\hline
			\small{Method} & \small{Standard VAE} & \small{IWAE-10} & \small{IWAE-20} & \small{IWAE-40} & \small{UHA-$10$} & \small{UHA-$20$} &  \small{RVRS-$0.1$} & \small{RVRS-$0.05$} & \small{RVRS-$0.025$}  \\ \hline\hline 
			
			%\small{Train -ELBO}  & \small{$92.00 \pm 0.10$} & \small{$88.66 \pm 0.08$} & \small{$87.93 \pm 0.03$} & \small{$87.35 \pm 0.08$} & \small{$87.01 \pm 0.08$} & \small{$86.04 \pm 0.33$} & \small{$87.58 \pm 0.08$} & \small{$87.10 \pm 0.07$} & \small{$86.87 \pm 0.06$} \\ \hline
			
			\small{Train $-$ELBO}  & \small{$92.00 \pm 0.10$} & \small{$88.66 \pm 0.08$} & \small{$87.93 \pm 0.03$} & \small{$87.35 \pm 0.08$} & \small{$87.01 \pm 0.08$} & \small{$86.04 \pm 0.33$} & \small{$87.58 \pm 0.08$} & \small{$87.10 \pm 0.07$} & \small{$86.87 \pm 0.06$} \\ \hline

			%\small{Test -ELBO}  & \small{$95.30 \pm 0.14$} & \small{$91.21 \pm 0.07$} & \small{$90.44 \pm 0.07$} & \small{$89.81 \pm 0.09$} & \small{$89.75 \pm 0.09$} & \small{$88.46 \pm 0.22$} & \small{$90.74 \pm 0.16$} & \small{$90.00 \pm 0.12$} & \small{$89.55 \pm 0.12$} \\ \hline
			
			\small{Test $-$ELBO} & \small{$95.30 \pm 0.14$} & \small{$91.21 \pm 0.07$} & \small{$90.44 \pm 0.07$} & \small{$89.81 \pm 0.09$} & \small{$89.75 \pm 0.09$} & \small{$88.46 \pm 0.22$} & \small{$90.74 \pm 0.16$} & \small{$90.00 \pm 0.12$} & \small{$89.55 \pm 0.12$} \\ \hline
			
			\small{ms / grad} & \small{$0.70$} & \small{$1.06$} & \small{$1.49$} & \small{$1.97$} & \small{$5.17$} & \small{$9.73$} &  \small{$1.19$}&  \small{$1.36$} & \small{$1.75$} \\ \hline
		\end{tabular}
	} % end resize
	\vspace{2mm}
	\caption{ 
        We report negative ELBO objectives (lower is better; mean $\pm$ standard deviation over $5$ replicates)
        computed on training data and held-out test data
        together with gradient step times for the VAE experiment in \secref{sec:vae}.
        Results obtained with a RTX 2070 GPU.
        This is the same table as in Table~\ref{table:vae} but includes objectives computed
        on the training set.
        }
	\label{table:vaebig}
\end{table}

\end{document}